\documentclass{article}

\usepackage{microtype}
\usepackage{graphicx}
\usepackage{subfigure}
\usepackage{booktabs} 
\usepackage[inline]{enumitem}
\usepackage{hyperref}

\usepackage[accepted]{mlsys2023}

\usepackage{tikz}
\usepackage{amsmath}

\usepackage{microtype}
\usepackage{graphicx}
\usepackage{subfigure}
\usepackage{booktabs} 
\usepackage{amsmath}
\usepackage[inline]{enumitem}
\usepackage{amsfonts}
\usepackage{tikz}
\usepackage{multirow}
\usepackage{makecell}
\usepackage[inline]{enumitem}
\usepackage{hyperref}
\usepackage[noend]{algpseudocode}
\usepackage{footmisc}

\newcommand*\circled[1]{\tikz[baseline=(char.base)]{
            \node[shape=circle,draw,inner sep=2pt] (char) {#1};}}
            
\usepackage{xspace}
\newcommand{\ea}{{et al.}\xspace}

\newcommand{\figcap}[1]{\small{\textbf{#1}}}

\newcommand{\name}{{Ada}\xspace}

\newcommand{\znote}[1]{\textit{\textcolor{red}{[zzy]: #1}}}

\newcommand{\TODO}[1]{\textcolor[HTML]{e41a1c}{{(#1)}}}

\usepackage{titling}
\usepackage{titlesec}

\mlsystitlerunning{Scaling up Decentralized Learning via Whitebox Analyses}

\begin{document}
\twocolumn[
\mlsystitle{Scaling up Data Parallelism in Decentralized Deep Learning}



\mlsyssetsymbol{equal}{*}
\begin{mlsysauthorlist}
\mlsysauthor{Bing Xie}{equal,meta}
\mlsysauthor{Junqi Yin}{equal,to}
\mlsysauthor{Zhenyu Zhou}{equal,goo}
\mlsysauthor{Sarp Oral}{to}
\mlsysauthor{Feiyi Wang}{to}
\end{mlsysauthorlist}

\mlsysaffiliation{to}{Oak Ridge National Laboratory, Tennessee, USA}
\mlsysaffiliation{goo}{Google Inc, CA, USA}
\mlsysaffiliation{meta}{Meta Platforms Inc, CA, USA}
\mlsyskeywords{Machine Learning, MLSys}
\vskip 0.3in

\begin{abstract}
 Although it has been extensively explored in theory, 
 decentralized learning is not yet green-lighted for production use, largely due to a lack of stability, scalability, and generality in large scale DNN training. 
To shed light on the production use of decentralized learning,  
this work studies decentralized data parallel training at scale. 
To this end, we introduce a benchmarking framework, namely \textit{DBench}, to host both centralized and decentralized DNN training.  Building upon DBench, we introduce a benchmarking methodology to uncover the correlations between model accuracy and the variances of parameter tensors by varying communication graphs and training scales.
Based on the benchmarking results, we observe that, (1) Similar to centralized learning, decentralized data parallel training also presents the issues of scalability and generality when the training scales up; (2) The model accuracy of decentralized learning is correlated to the number of connections in a communication graph;  \textcolor{black}{(3) The model accuracy of decentralized learning is surprisingly sensitive to the variance of parameter tensors across model replicas}. 
Built upon the observations, 
 we propose {\it \name}, a decentralized adaptive approach that performs large scale DNN training following a decentralized SGD method and adapting the communication graph in use dynamically throughout training iterations.
We apply Ada on large scale training and observe that Ada can obtain the best convergence rates consistently in decentralized DNN training, and delivers equally or comparably good model accuracy for all sample applications as centralized learning does,  even when training ResNet50 for ImageNet-1K on the scale of 1008 GPUs. 
\end{abstract}
]



\section{Introduction}
\label{sec:intro}

The recent successes of Deep Learning (DL) have encouraged continued investment across industries and domain sciences. In particular, ranging from the traditional AI (e.g., image processing, speech recognition), to pharmaceutical and biomedical sciences (e.g., drug discovery, protein-structure prediction), and 
to fusion, combustion and nuclear energy (e.g., disruption predictor, nuclear power plant), more and more applications are actively exploiting ever-larger deep neural networks (DNNs) for production use.

\textit{\textbf {Motivation}}. In distributed DNN training, data parallelism still holds a dominant position in support of large scale machine learning for its simplicity, efficiency and scalability.  In a typical data parallel training run, an application  creates multiple replicas of a DNN model and distributes the model replicas among a group of accelerators (e.g., CPUs, GPUs, TPUs)\footnote{\textcolor{black}{In this work, our implementation focuses on GPUs.}}, where each accelerator processes a different shard of training data throughout a number of \textit{iterations}.  Over the course of an  iteration, each accelerator performs forward and backward passes of computations independently, but synchronizes the tensors (multi-dimensional arrays) of gradients or parameters among the accelerators before applying weight updates. 
Beyond the successes of data parallel training alone, 
data parallelism in decentralized learning
has gained an increasing attention for its theoretically proved good model accuracy, high communication efficiency, and 
the corresponding cheap training cost. In particular, centralized data parallel training, sometimes referred to as synchronous DL
training, averages gradient tensors across all accelerators and accordingly maintains a \textit{globally consistent state} for the hosted model replicas. Comparatively,    
decentralized learning \cite{hogwild, lian2017can, koloskova2019decentralized}, sometimes referred to as gossip learning or asynchronous DL training, averages \textit{the tensors of parameters} 
among accelerators \textit{locally} based on some predefined \textit{communication graphs}. In decentralized learning, accelerators are structured into a communication graph such like a ring, or a torus, based on which each accelerator averages parameter tensors with its neighbor accelerators defined in the graph (discussed in \S\ref{subsec:learning_definition}). Clearly, decentralized learning maintains \textit{locally consistent states} of the model replicas among the graph-defined neighbors. 

\textit{\textbf{Limitations of the state-of-the-art approaches}}. In the context of decentralized data parallel training, on one hand, the existing efforts centered on advancing the algorithms, such as decentralized optimization \cite{koloskova2019decentralized, koloskova2020decentralized, elgabli2020q, lian2018asynchronous, reisizadeh2019exact}, static and dynamic communication graphs \cite{mai2020kungfu,nvidiaoptimized,sergeev2018horovod}, and distributed average consensus \cite{xiao2004, koloskova2019decentralized}, building the theoretical foundation for decentralized learning. On the other hand, although it is widely considered as a communication-efficient alternative to centralized learning, decentralized learning has not yet been vetted for production use largely due to a lack of stability caused by the many fluctuations and 
vast differences in losses and/or accuracy of local training results across accelerators when scaling up. The previous studies \cite{ddp} believe that, these fluctuations and differences may be  deleterious 
to model convergence, especially when scaling up accelerators, datasets, and/or models. 

\textit{\textbf{Our proposals}}. 
For decentralized learning, to bridge the gap between its theories and its production use in large scale DNN training, 
this work discusses the scalability issue through uncovering the decentralized learning internals. We introduce a benchmarking framework, namely \textit{DBench}, to profile training runs of centralized/decentralized learning via enabling two configurable parameters: communication graph and training scale. Moreover, we collect and analyze the model accuracy of training/testing data and the local results on L2-norm of parameter tensors per GPU before averaging parameters. 
Utilizing DBench, we introduce a benchmarking methodology to conduct a group of controlled experiments on centralized and decentralized learning with various configurations on communication graph and training scale. Built upon the benchmarking results, we conduct analysis to uncover the correlations between model accuracy and variances of parameter tensors across model replicas by varying communication graphs and training scales. 
Built upon the analysis, we propose {\it \name}, a decentralized adaptive approach that performs DNN training following a decentralized stochastic gradient descent (SGD) method and adapts communication graphs throughout training iterations. 

\textit{\textbf{Key insights and contributions}}.  To the best of our knowledge, our proposals of DBench and Ada separately on the white-box analysis and decentralized adaptive approach are the first efforts on understanding and improving the scalability and generality of decentralized learning at scale and for production use. 
We summarize our key insights and contributions as follows. 
\begin{enumerate}
\item We implement DBench as a benchmarking framework to host centralized and decentralized DNN training. Built on DBench, we introduce a benchmarking methodology and conduct a group of controlled experiments to understand the characteristics of decentralized learning and uncover the correlations between model accuracy and variances of parameter tensors across model replicas in a training run. \textcolor{black}{ We publish the benchmarking results\footnote{DBench: \url{https://anonymous.4open.science/r/dbench\_figs-08CB/} \label{foot}} and plan to open-source the Ada software upon the acceptance}.
 
\item Built upon the white-box analysis, we find that, (1) Similar to centralized learning, decentralized learning also presents the scalability and generality issues when scaling up. (2) For decentralized learning, model accuracy is positively correlated to the number of connections in communication graphs: the more connections a graph has, the better model accuracy the graph-guided training run can obtain. 
  (3) The conventional learning-rate configurations, widely used in centralized DL training, do not always work well in decentralized DL training, especially when training on larger scales and/or with more connections in  communication graphs. (4) \textcolor{black}{New to the existing decentralized learning studies, we find that the degree of parameter-tensor variance is correlated to the number of connections of a communication graph and to the model accuracy, especially at the early stage of a training run.} (5) For decentralized DL training, an adaptive
approach with varying communication graphs might be more beneficial than a static solution with
a fixed graph in the context of accuracy and training cost.

\item Motivated by the findings, we propose \name, a decentralized adaptive approach that trains DNN models with a decentralized SGD method by varying communication graphs dynamically across iterations. We experimented Ada on four sample applications at large scales and observed that our approach delivers equally or comparably good accuracy as centralized learning does for all applications, even when training ResNet50 for ImageNet-1K on the scale of 1008 GPUs. We will release Ada soon. 
 \end{enumerate}

\section{Background}
\label{sec:background}

\subsection{Data Parallelism in Distributed DNN Training}
\label{subsec:learning_definition}

For data parallelism in distributed DL training, a typical training run executes an iterative learning algorithm such as SGD and Laryngeal Mask Airway (LMA), 
among a number of GPUs, during which each GPU works on a replica and the same set of parameters of a DNN model. 
For such a DL training run in both centralized (synchronous) and decentralized (asynchronous) learning, 
the training data is partitioned into one or more equal-sized {\it batches}, each of which is processed on a different GPU. After the run starts, the GPUs independently process computations on the model replicas they host throughout a number of \textit{iterations} and 
in particular, in the course of \textcolor{black}{an iteration},  each GPU updates parameters of a model replica by the following computational procedure: 
\begin {enumerate*}
\item The forward pass to compute loss.
\item The backward pass to compute gradients of the parameters.
\item The optimization step to update the parameters.
\end{enumerate*}

Particularly, in an iteration of a centralized training run, before updating parameters, the gradient tensors between GPUs are averaged globally by executing collective operators (e.g., \texttt{AllReduce}) to ensure model replicas are updated identically based on the globally averaged gradient tensors and accordingly remain in a \textit{globally consistent state}.
Comparatively, for a decentralized training run, parameters are updated based on a communication graph that is predefined before the run starts. 
In particular, to update parameters in an iteration, decentralized learning takes two steps:
\begin{enumerate*}
\item each GPU updates the parameters locally using the local gradient tensors computed in the backward pass.
\item each GPU averages its locally updated parameter tensors among the graph-defined neighbor GPUs.
\end{enumerate*}
Clearly, decentralized learning ensures \textit{locally consistent states} of model replicas among the neighbors based on communication graphs.

Shown as Figure~\ref{fig:top}, in decentralized learning, a typical communication graph can be a ring, a torus, etc. 
For different graphs, each GPU in the system averages parameter tensors with a different number of neighbors. For example, for a ring-, or torus-based run, each GPU averages with two or four neighbors, respectively. It is worth noting that, for a decentralized training run with a complete graph, each GPU synchronizes with every other GPU in the system and accordingly maintains a globally consistent state for model replicas. Compared to averaging in centralized DL training, complete-graph based decentralized training is different as it averages parameter tensors. Centralized learning, in another word, averages gradient tensors based on complete graphs.

\def\ab{.4}
\tikzset{
  net node/.style = {circle, minimum width=1*\ab cm, inner sep=0pt, outer sep=0pt, ball color=blue!50!cyan},
  self node/.style = {circle, minimum width=1*\ab cm, inner sep=0pt, outer sep=0pt, ball color=red},
  neighbor node/.style = {circle, minimum width=1*\ab cm, inner sep=0pt, outer sep=0pt, ball color=green},
  net connect/.style = {line width=1pt, draw=blue!50!cyan!25!black},
  net thick connect/.style = {net connect, line width=2.5pt, draw=red},
  net directed connect/.style = {->, line width=1pt, draw=blue!50!cyan!25!black},
  net thick directed connect/.style = {->, net connect, line width=2.5pt, draw=red},
}

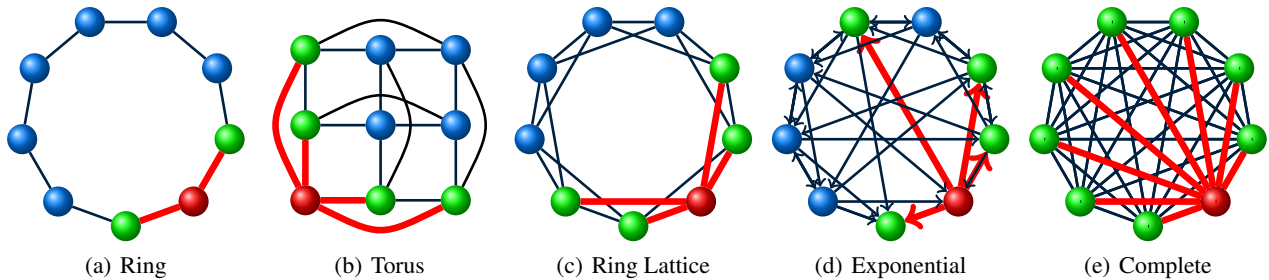
\begin{figure*}[t!]
\centering
\subfigure[Ring]{
\begin{tikzpicture}[scale=0.7]
  \path (-90+0*40:2) node (n0) [neighbor node] {};
  \path (-90+1*40:2) node (n1) [self node] {};
  \path (-90+2*40:2) node (n2) [neighbor node] {};
  \path (-90+3*40:2) node (n3) [net node] {};
  \path (-90+4*40:2) node (n4) [net node] {};
  \path (-90+5*40:2) node (n5) [net node] {};
  \path (-90+6*40:2) node (n6) [net node] {};
  \path (-90+7*40:2) node (n7) [net node] {};
  \path (-90+8*40:2) node (n8) [net node] {};
  
  \path [net thick connect] (n0) -- (n1) -- (n2);
  \path [net connect] (n2) -- (n3) -- (n4) -- (n5) -- (n6) -- (n7) -- (n8) -- (n0);
\end{tikzpicture}
}
\subfigure[Torus]{
  \begin{tikzpicture}[scale=0.5,line width=1pt]
   \path (0, 0) node (n0) [self node] {};
   \path (2, 0) node (n1) [neighbor node] {};
   \path (4, 0) node (n2) [neighbor node] {};
   \path (0, 2) node (n3) [neighbor node] {};
   \path (2, 2) node (n4) [net node] {};
   \path (4, 2) node (n5) [net node] {};
   \path (0, 4) node (n6) [neighbor node] {};
   \path (2, 4) node (n7) [net node] {};
   \path (4, 4) node (n8) [net node] {};
   
  \path [net connect] (n2) -- (n1) -- (n0);
  \path [net connect] (n5) -- (n4) -- (n3);
  \path [net connect] (n8) -- (n7) -- (n6);
  \path [net connect] (n0) -- (n3) -- (n6);
  \path [net connect] (n1) -- (n4) -- (n7);
  \path [net connect] (n2) -- (n5) -- (n8);
  
  \path [net thick connect] (n0) -- (n1);
  \path [net thick connect] (n0) -- (n3);
  
   \draw[line width=2.5pt,draw=red] (n0) .. controls (2, -1) .. (n2);
   \draw (n3) .. controls (2, 3) .. (n5);
   \draw (n6) .. controls (2, 5) .. (n8);
   \draw[line width=2.5pt,draw=red] (n0) .. controls (-1, 2) .. (n6);
   \draw (n2) .. controls (5, 2) .. (n8);
   
   \draw (n1) .. controls (3, 2) .. (n7);
  \end{tikzpicture}
}
\subfigure[Ring Lattice]{
\begin{tikzpicture}[scale=0.7]
  \path (-90+0*40:2) node (n0) [neighbor node] {};
  \path (-90+1*40:2) node (n1) [self node] {};
  \path (-90+2*40:2) node (n2) [neighbor node] {};
  \path (-90+3*40:2) node (n3) [neighbor node] {};
  \path (-90+4*40:2) node (n4) [net node] {};
  \path (-90+5*40:2) node (n5) [net node] {};
  \path (-90+6*40:2) node (n6) [net node] {};
  \path (-90+7*40:2) node (n7) [net node] {};
  \path (-90+8*40:2) node (n8) [neighbor node] {};
  
  \path [net thick connect] (n0) -- (n1) -- (n2);
  \path [net connect] (n2) -- (n3) -- (n4) -- (n5) -- (n6) -- (n7) -- (n8) -- (n0);
  
  \path [net connect] (n0) -- (n2);
  \path [net connect] (n2) -- (n4);
  \path [net connect] (n3) -- (n5);
  \path [net connect] (n4) -- (n6);
  \path [net connect] (n5) -- (n7);
  \path [net connect] (n6) -- (n8);
  \path [net connect] (n7) -- (n0);
  
  \path [net thick connect] (n1) -- (n3);
  \path [net thick connect] (n8) -- (n1);
\end{tikzpicture}
}
\subfigure[Exponential]{
\begin{tikzpicture}[scale=0.7]
  \path (-90+0*40:2) node (n0) [neighbor node] {};
  \path (-90+1*40:2) node (n1) [self node] {};
  \path (-90+2*40:2) node (n2) [neighbor node] {};
  \path (-90+3*40:2) node (n3) [neighbor node] {};
  \path (-90+4*40:2) node (n4) [net node] {};
  \path (-90+5*40:2) node (n5) [neighbor node] {};
  \path (-90+6*40:2) node (n6) [net node] {};
  \path (-90+7*40:2) node (n7) [net node] {};
  \path (-90+8*40:2) node (n8) [net node] {};
  
  \path [net connect] (n2) -- (n3) -- (n4) -- (n5) -- (n6) -- (n7) -- (n8) -- (n0);
  
  \path [net thick directed connect] (n1) -- (n2);
  \path [net thick directed connect] (n1) -- (n3);
  \path [net thick directed connect] (n1) -- (n5);
  \path [net thick directed connect] (n1) -- (n0);
  
  \path [net directed connect] (n2) -- (n3);
  \path [net directed connect] (n2) -- (n4);
  \path [net directed connect] (n2) -- (n6);
  \path [net directed connect] (n2) -- (n1);
  
  \path [net directed connect] (n3) -- (n4);
  \path [net directed connect] (n3) -- (n5);
  \path [net directed connect] (n3) -- (n7);
  \path [net directed connect] (n3) -- (n2);
  
  \path [net directed connect] (n4) -- (n5);
  \path [net directed connect] (n4) -- (n6);
  \path [net directed connect] (n4) -- (n8);
  \path [net directed connect] (n4) -- (n3);
  
  \path [net directed connect] (n5) -- (n6);
  \path [net directed connect] (n5) -- (n7);
  \path [net directed connect] (n5) -- (n0);
  \path [net directed connect] (n5) -- (n4);
  
  \path [net directed connect] (n6) -- (n7);
  \path [net directed connect] (n6) -- (n8);
  \path [net directed connect] (n6) -- (n1);
  \path [net directed connect] (n6) -- (n5);
  
  \path [net directed connect] (n7) -- (n8);
  \path [net directed connect] (n7) -- (n0);
  \path [net directed connect] (n7) -- (n2);
  \path [net directed connect] (n7) -- (n6);
  
  \path [net directed connect] (n8) -- (n0);
  \path [net directed connect] (n8) -- (n1);
  \path [net directed connect] (n8) -- (n3);
  \path [net directed connect] (n8) -- (n7);
  
\end{tikzpicture}
}
\subfigure[Complete]{
\begin{tikzpicture}[scale=0.7]
  \path (-90+0*40:2) node (n0) [neighbor node] {};
  \path (-90+1*40:2) node (n1) [self node] {};
  \path (-90+2*40:2) node (n2) [neighbor node] {};
  \path (-90+3*40:2) node (n3) [neighbor node] {};
  \path (-90+4*40:2) node (n4) [neighbor node] {};
  \path (-90+5*40:2) node (n5) [neighbor node] {};
  \path (-90+6*40:2) node (n6) [neighbor node] {};
  \path (-90+7*40:2) node (n7) [neighbor node] {};
  \path (-90+8*40:2) node (n8) [neighbor node] {};
  
  \foreach \i in {0,...,8}
    \foreach \j in {0,...,8}
      \path [net connect]
        (n\i) -- (n\j);
        
  \foreach \i in {0,...,8}
      \path [net thick connect]
        (n\i) -- (n1);
\end{tikzpicture}
}

\caption{\figcap{Representative communication graphs used in decentralized DNN training.} In each communication graph, we use a red node and red arrows/lines to denote a node and its  connections (neighbors) in the graph.  }
\label{fig:top}
\end{figure*}

\subsection{Theory of Decentralized Learning Training}
\label{subsec:theory}

\begin{table*}[!thbp]
\center
\begin{tabular}{c c c c c}
\toprule
\textbf{Graph Name} & \makecell[c]{\textbf{Number of Neighbors} \\ \textbf{(Node Degree)}} & \textbf{Number of Edges} & \textbf{Symmetry} & \textbf{Directed} \\
\toprule
Ring & $2$ & $n$ & Axial and central & No \\
\midrule
Torus & $4$ & $2n$ & Axial and central & No \\
\midrule
Ring Lattice & $2k$ & $kn$ & Axial and central & No \\
\midrule
Exponential & $\lfloor\log_{2} (n-1)\rfloor + 1$ & $n(\lfloor\log_{2} (n-1)\rfloor + 1)$ & Central & Yes \\
\midrule
Complete & $n-1$ & $\frac{n(n-1)}{2}$ & Axial and central & No \\

\bottomrule
\end{tabular}
\caption{\figcap{Characteristics of the representative graphs used in decentralized DNN training.} $n$ denotes the number of nodes in a graph. For ring lattice, there is a new parameter $k$, called coordination number, which controls the number of neighbors a node is connected to (discussed in Section~\ref{sec:solution:design_impl}). \iffalse This work focuses on understanding regular graphs where each node is connected to the same number of neighbors (node degree).\fi}\label{tab:graph}
\end{table*}

Assuming this scenario, before a decentralized training run starts, there exist $n$ GPUs in the system and each GPU hosting an identical DNN model replica. Formally speaking, for such a training run, its training scheme is executed based on a communication graph $G\!=\!\langle V, E \rangle$, where $V\!=\!\{v_1, v_2, \cdots, v_n\}$ represents the $n$ GPUs (or called as nodes in the view of a communication graph) in use and $E \in \mathbb{R}^{n\times n}$ represents the adjacency matrix of $G$. Here, the element $E_{ij}$ of $E$ indicates $v_i$ and $v_j$ are adjacent (neighbors to each other with $E_{ij}>0$) 
or not ($E_{ij}=0$) in the graph. 
In production use, $E$ is usually a symmetric matrix and $G$ is usually a symmetric and regular graph, where each node in $G$ is connected to the same number of neighbors.

This work mainly focuses on five representative communication graphs. We present the graphs as Figure~\ref{fig:top} and \textcolor{black}{summarize the characteristics of these five graphs in Table~\ref{tab:graph}}. Among these graphs, we include exponential graphs to enrich our study. 
In particular, an exponential graph is a typical example of expander graphs, which are not extensively explored in production use but are well studied in decentralized learning theories with theoretically guaranteed model accuracy \cite{ying2021exponential}.  We discuss the performance of exponential graphs in detail in Section \ref{sec:analysis} and refer the interested readers to Appendix~\ref{sec:spectral} for the introduction of expander graphs.

Beyond built on communication graphs, decentralized learning aims to optimize the averaged object function among the model replicas across GPUs:
$$\min\limits_{\mathbf{\theta}\in\mathbb{R}^N}\frac{1}{n}\sum\limits_{i=1}^n\mathop{\mathbb{E}}_{d_i\sim\mathcal{D}}F_i(\mathbf{\theta}, d_i)$$

where $\mathbf{\theta}$ is a vector representing model parameters and $d_i$ is a data sample (\textcolor{black}{a batch}) from a data 
distribution $\mathcal{D}$ (\textcolor{black}{the training dataset}) on the $i$-th GPU.

To achieve this objective, a decentralized SGD~\cite{lian2017can} works as the optimization step (\S\ref{subsec:learning_definition}) that updates the parameters  in each iteration of the training run. In particular, once the local gradients $\nabla F_i(\mathbf{\theta}_i, d_i)$ are computed in the backward pass, each GPU updates its local model parameters by averaging the parameters among its neighbors defined in the graph,
following the formula: $\sum\limits_{j=1}^{n}E_{ij}\mathbf{\theta}_j$. Next, the updated parameters per GPU are further optimized by decreasing with the local gradient, following the formula: $\gamma\nabla F_i(\mathbf{\theta}_i, d_i)$.
Here, $\gamma$ is the learning rate. In summary, the decentralized SGD scheme expects the parameters of each GPU ($\mathbf{\theta}_i$) to converge and the trained model takes the parameters $\mathbf{\theta}$ as the average over all $\mathbf{\theta}_i$ at the end.
It is worth noting that, this optimization procedure is in a reversed order as we presented in Section~\ref{subsec:learning_definition}, and is proved theoretically as no effect to model convergence \cite{lian2017can}.

\section{DBench and White-box Analysis}
\label{sec:analysis}

This section presents a white-box analysis on decentralized DL training built on a benchmarking framework, namely  \textit{DBench}\footnote{Due to the space limitation, we present part of the result figures in this section and the supplementary materials, and leave all figures on an anonymous repository\textsuperscript{\ref{foot}}. }.
Furthermore, based on the observations derived from the benchmarking analysis, we propose a decentralized adaptive training approach, namely {\it Ada}  (\S\ref{sec:solution}). \textcolor{black}{Our goal is to obtain consistent observations and findings from small-scale settings in DBench for low benchmarking cost, and build and experiment Ada on large-scale settings to verify its generality and scalability. }

 To summarize, we introduce \textit{DBench} to host the benchmarking experiments of DNN training runs for both centralized and decentralized learning. To this end, we enable two configurable parameters in DBench, including communication graph and training scale. We collect the profiling data on model accuracy for both training and testing data and on the local results of the L2-norm of parameter tensors per GPU before averaging (discussed in \S\ref{subsec:learning_definition}). Built upon DBench, we propose a benchmarking methodology and design a group of controlled experiments on four sample applications (see Table~\ref{tab:exp}) with various configurations on communication graph and training scale (\S\ref{subsec:methodology}). Based on the benchmarking results, we conduct a white-box analysis upon model accuracy (\S\ref{subsec:accuracy}) and variances of parameter tensors across GPUs (\S\ref{subsec:variance}). 

\subsection{Benchmarking Methodology}
\label{subsec:methodology}

\begin{table*}[!t]
\center
\begin{tabular}{c c c c c}
\toprule
\textbf{Task} & \makecell{\textbf{Model}\\ \textbf{(\#Parameters)}} & \textbf{Dataset} & \makecell{\textbf{Batch Size}\\ \textbf{Per GPU}}    & \makecell{\textbf{Learning Rate}\\ \textbf{Scheduling}}  \\
\midrule
& \makecell{ResNet20\\(0.27M)} & CIFAR10 & 128 & \makecell{All: one cycle \\ epoch={[}(1,23), (23,46), (46,300) {]}\\ lr={[}(0.15, 3s), (3s, 0.15s),(0.15s, 0.015s){]}\\
s=1(complete/ring/torus)\\
s=k(ours)=max(\#GPUs//9-epoch//50, 2)} \\
\cmidrule(l){2-5}
\makecell{Image\\ Classification} & \makecell{ResNet50\\(25.56M)} & Imagenet-1k & \makecell{32 (small)\\ 16 (large)} & \makecell{complete/ring/torus/exponential: warmup, multi-step\\ epoch={[}(0, 5), (5, 30), (30, 60), (60, 80), (80, 90){]}\\ lr = {[}(lr$_0$,lr$_0$s),(lr$_0$s,lr$_0$s),($\frac{lr_0}{10}$s,$\frac{lr_0}{10}$s),($\frac{lr_0}{100}$s,$\frac{lr_0}{100}$s){]}\\ lr$_0$=0.1 (complete/ring/torus), 0.8(exponential)\\ ours: warmup, multi-step\\ epoch={[}(0, 5), (5, 20), (20, 40), (40, 60), (60, 90){]}\\ lr={[}(0.1,0.1s),(0.1s,0.1s),(0.05s,0.05s),(0.025s,0.025s){]}\\ s=Batch\_Size*(k+1)/256\\ k=2(ring),4(torus),6(exponential),\#GPU-1(complete)\\ k(ours)=max(\#GPUs//9-epoch, 2)} \\
\cmidrule(l){2-5}
& \makecell{DenseNet100\\(4.07M)} & CIFAR10     & 128 & \makecell{All: one cycle\\ epoch={[}(1,23),(23,46),(46,300){]}\\ lr={[}(0.15,3s),(3s,0.15s),(0.15s,0.015s){]}\\
s=1(complete/ring/torus)\\
s=k(ours)=max(\#GPUs//9-epoch//50, 2)} \\

\midrule
\makecell{Language \\ Modeling} & \makecell{LSTM\\(28.95M)}  & WikiText2 & 32 & \makecell{All: warmup, multi-step \\ epoch={[}(0, 5), (5, 150), (150, 225), (225, 300){]}\\ lr={[}(2.5, 2.5s), (2.5s, 2.5s), (0.25s, 0.25s), (0.025s, 0.025s){]}\\ s=Batch\_Size*(k+1)/24\\ k=2(ring),4(torus),\#GPU-1(complete)\\k(ours)=max(\#GPUs//9-epoch//50, 2) } \\

\bottomrule
\end{tabular}
\caption{\figcap{Models, datasets, and training parameters. 
}}\label{tab:exp}
\end{table*}

\subsubsection{System Settings}
\label{subsec:environment}

\textbf{Hardware.} We performed all experiments on the Summit supercomputer housed at the Oak Ridge Leadership Computing Facility (OLCF). 
Summit is a 148.6 petaFLOPS (double precision) IBM-built supercomputer, consisting of 4,608  AC922 compute nodes with each node equipped with 2 IBM POWER9 CPUs and 6 NVIDIA V100 GPUs. 
 Summit is considered as ideally suited for Deep Learning workloads, due to its node-local NVMe (called burst buffer) and Tensor Cores on V100 for faster low-precision operations. 
 Moreover, its NVLink 2.0 and EDR InfiniBand interconnect provides 50 GB/s and 23 GB/s peak network bandwidths for intra-node and inter-node communication.

\textbf{Models and datasets.} We conduct the white-box analysis on four applications, covering the popular AI/ML tasks in both image classification  and natural language processing (NLP), shown in details in Table~\ref{tab:exp}. 
{\color{black}We set the batch size per GPU and learning rate in the analysis by following the policies introduced in  \cite{koloskova2019decentralized}, where the learning-rate setting is conventional. We configure the learning rate linearly scaled to two factors: global batch size and the connection degree of a communication graph. }
We expect these applications can present a good coverage of the DNN model scales ranging from small to medium (0.27M parameters to 28.95M parameters) models.
 Moreover, we expect these applications can be representative for a large number of deep learning applications in production use. 


\subsubsection{DBench and Benchmarking Experiments}

\begin{table}[!t]
\centering
\scriptsize
\begin{tabular}{c c c c c}
\toprule
\textbf{Application} & \makecell{\textbf{\#Parameter}\\ \textbf{Tensors}} & \makecell[c]{\textbf{Tensor Size}\\ \textbf{[min, max]}} & \textbf{Topology} & \textbf{\#GPUs} \\
\midrule
\makecell{ResNet20\\ CIFAR10} & 65 & [10, 36864] & \multirow{4}*[-3ex]{\makecell{centralized \\ complete,\\ decentralized \\ ring,\\ torus, \\ exponential, \\ complete}} & \multirow{4}*[-5ex]{\makecell{12,24,\\48,96}} \\
\cmidrule(l){1-3}
\makecell{ResNet50\\ ImageNet-1k} & 161 & [64, 2359296] & & \\
\cmidrule(l){1-3}
\makecell{DenseNet100\\ CIFAR10} & 299 & [10, 172872] & & \\
\cmidrule(l){1-3}
\makecell{LSTM\\ WikiText2} & 14 & [2600, 18764850]& & \\
\bottomrule
\end{tabular}
\caption{\figcap{Benchmarking experiment settings.}}\label{tab:bench}
\end{table}

We realize \textit{DBench} to investigate the scalability and generality of decentralized DNN training. 
DBench is built on a DL framework that supports decentralized SGD processes on PyTorch~\cite{koloskova2019decentralized}. 
 To obtain an in-depth understanding on the internals of DNN training, 
DBench collects the L2-norm of parameter and gradient tensors via the PyTorch interface,  \texttt{torch.tensor.norm()}. \textcolor{black}{We choose L2-norm as it is widely used and the default metric enabled in PyTorch}. Furthermore, our analysis focuses on parameter tensors as we see parameter tensors address the variances across GPUs more efficiently. 
Utilizing DBench, we conduct a group of benchmarking experiments. Here, an experiment performs a training application on a given number of GPUs (a training scale) with five SGD implementations. 

In particular, across scales and experiments, the five SGD implementations, built on PyTorch v1.7.0 and Spectrum MPI v10.4.0, include: 

\begin{enumerate}
\item Centralized learning, namely \textit{centralized complete} or \texttt{C\_complete}. It issues synchronized averaging on gradient tensors across all GPUs (based on a complete graph). We realize this centralized SGD implementation with PyTorch Distributed Data Parallel (DDP). In this study, the \texttt{C\_complete} runs serve as the baseline to evaluate the performance of various decentralized SGD implementations. 

\item Decentralized learning with complete graph, namely \textit{decentralized complete} or \texttt{D\_complete}. It issues synchronized averaging on parameter tensors across all GPUs on a complete graph at each iteration.  

\item Decentralized learning with ring, namely \textit{decentralized ring} or \texttt{D\_ring}. This SGD implementation averages parameter tensors among the neighbors based on a ring. 

\item Decentralized learning with torus, namely \textit{decentralized torus} or \texttt{D\_torus}. Similar to decentralized ring, this SGD implementation averages parameter tensors among the neighbors based on a torus. 

\item Decentralized learning with exponential graph, namely \textit{decentralized exponential} or \texttt{D\_exponential}. Similarly, this SGD implementation averages parameter tensors among the neighbors based on an exponential graph. Particularly, for the $i$th node in an exponential graph with overall $n$ nodes, the neighbors $S_i= \{(i+2^{m}) \% n\}$, where $m=0, 1, 2,...,\lfloor\log_{2} (n-1)\rfloor. $ 
\end{enumerate}

Beyond performing the above five SGD implementations in an experiment, we also vary the training scales per experiment and present the details of the varying parameters and their values we used in Table~\ref{tab:bench}. 

\subsection{Study on Model Accuracy}
\label{subsec:accuracy}

\begin{figure}[t]
    \centering
    \subfigure[ResNet50, \textit{D\_ring}]{\includegraphics[width=0.23\textwidth]{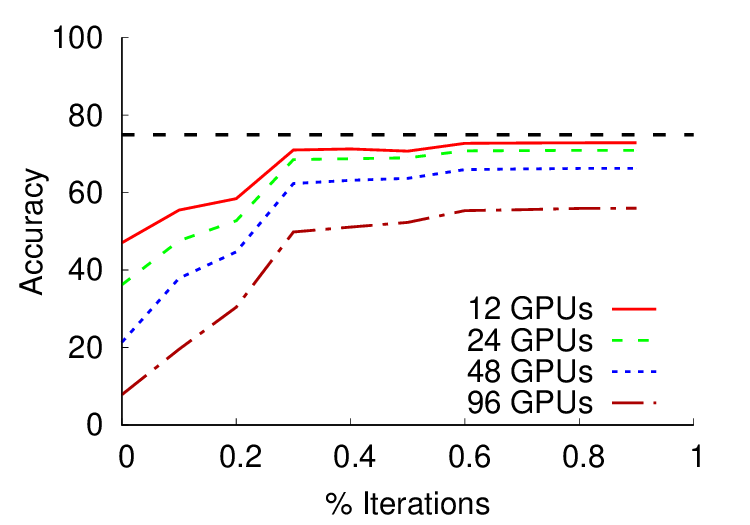}} 
    \subfigure[ResNet50, \textit{D\_complete}]{\includegraphics[width=0.23\textwidth]{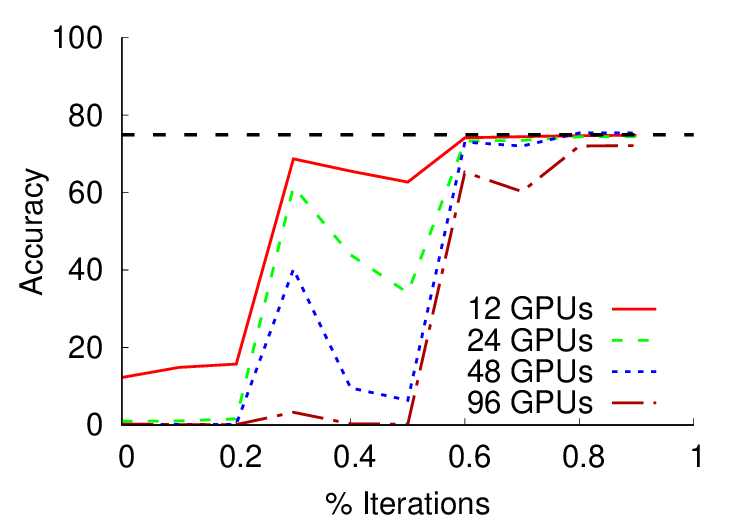}} 
    \caption{\figcap{Model accuracy of ResNet50 trained with decentralized ring (left) and decentralized complete (right).} Decentralized ring and complete are defined in \S\ref{subsec:learning_definition}.}
    \label{fig:t_accuracy}
\end{figure}

\begin{figure*}[h!]
    \centering
    
    \subfigure[ResNet20, 24 GPUs]{\includegraphics[width=0.245\textwidth]{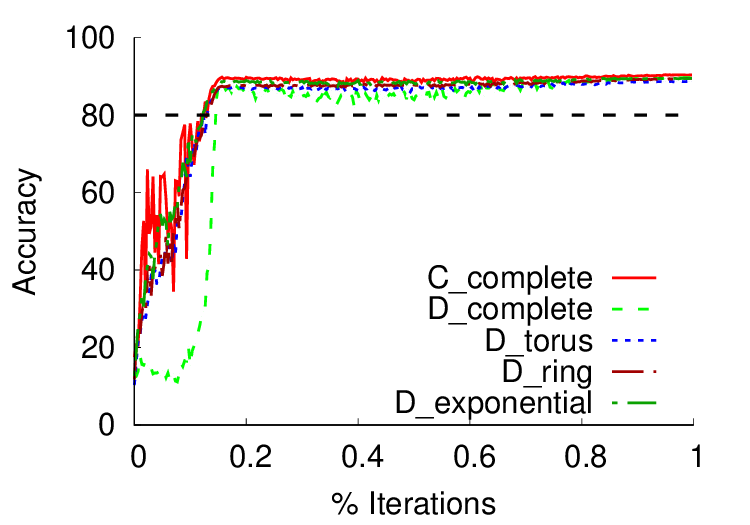}} 
    \subfigure[DenseNet100, 24 GPUs]{\includegraphics[width=0.245\textwidth]{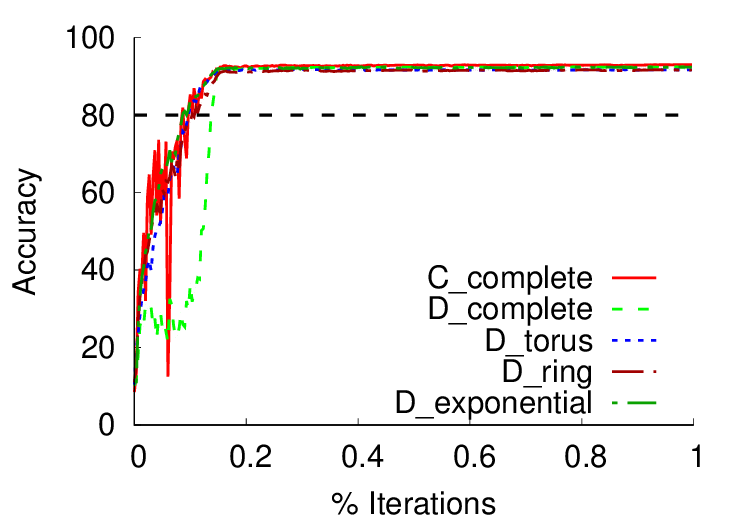}} 
    \subfigure[ResNet50, 24 GPUs]{\includegraphics[width=0.245\textwidth]{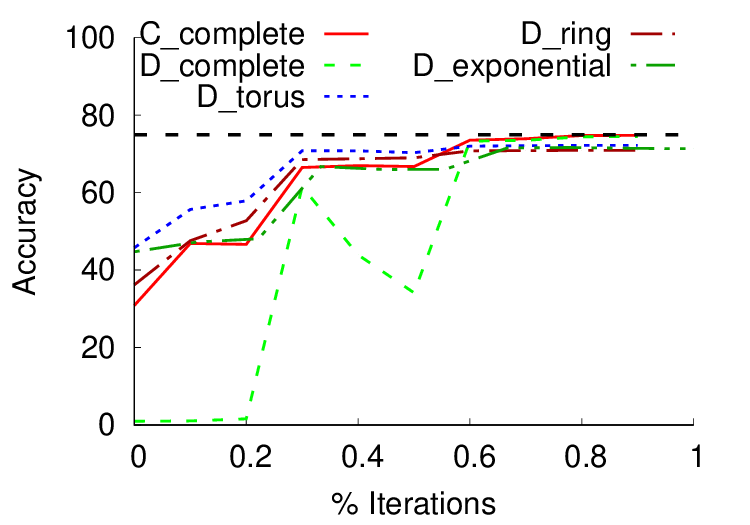}} 
    \subfigure[LSTM, 24GPUs]{\includegraphics[width=0.245\textwidth]{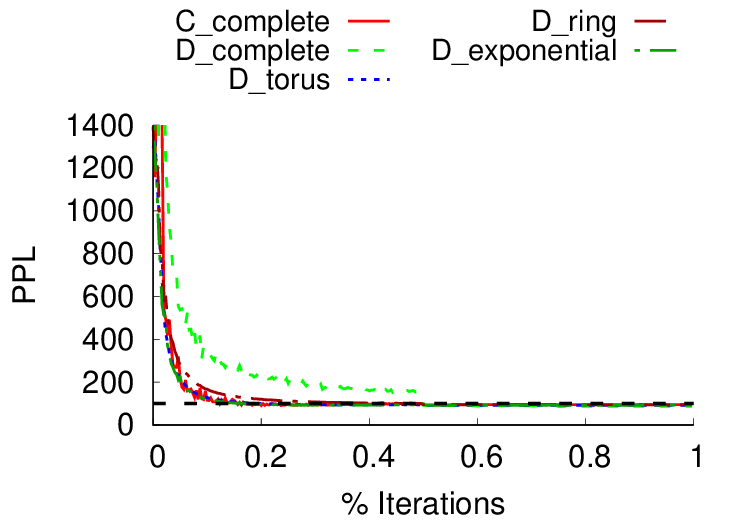}} 
   
   \subfigure[ResNet20, 48GPUs]{\includegraphics[width=0.245\textwidth]{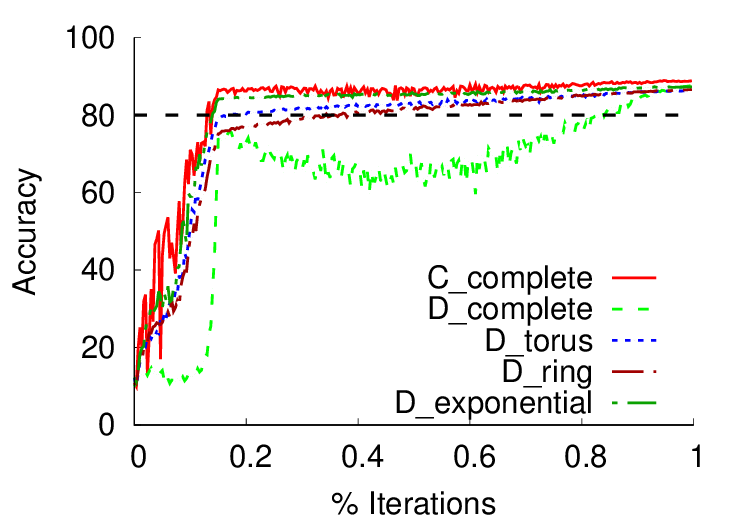}} 
    \subfigure[DenseNet100, 48GPUs]{\includegraphics[width=0.245\textwidth]{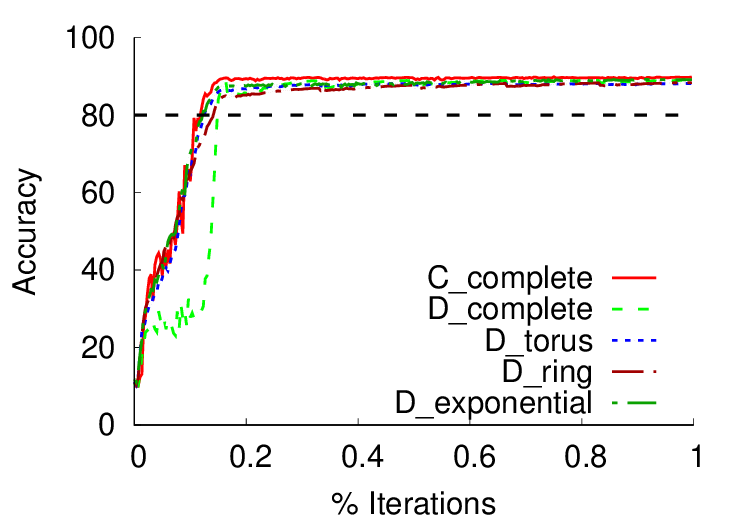}} 
    \subfigure[ResNet50, 48GPUs]{\includegraphics[width=0.245\textwidth]{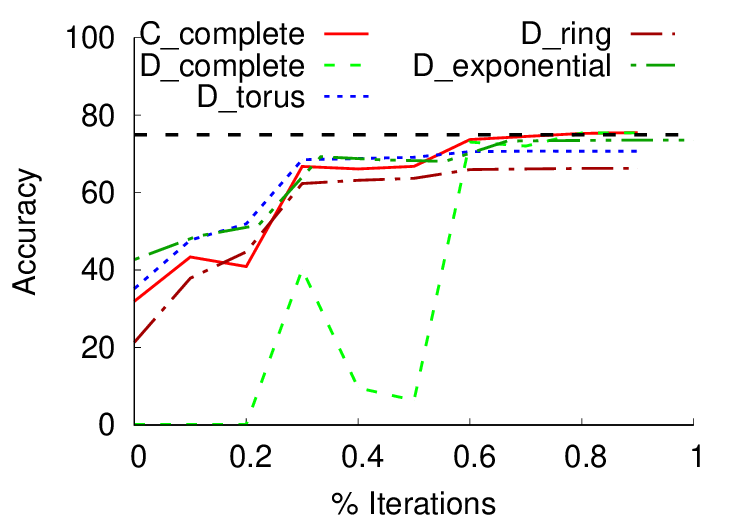}} 
    \subfigure[LSTM, 48GPUs \label{fig:b}]{\includegraphics[width=0.245\textwidth]{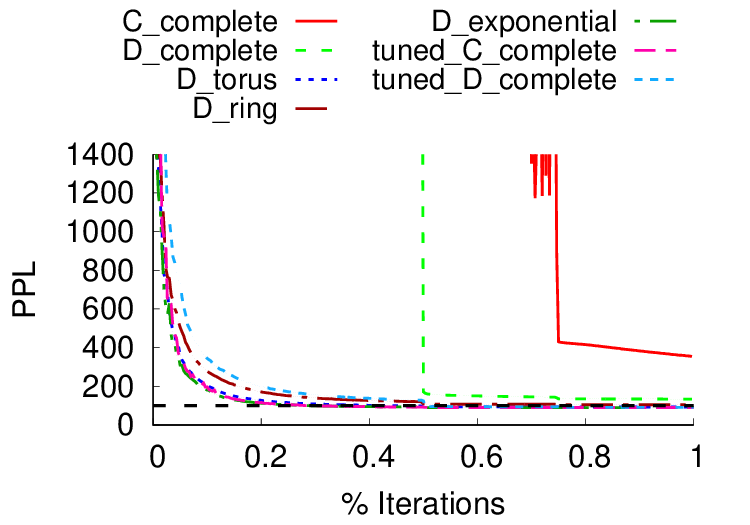}} 
     
    \subfigure[ResNet20, 96GPUs]{\includegraphics[width=0.245\textwidth]{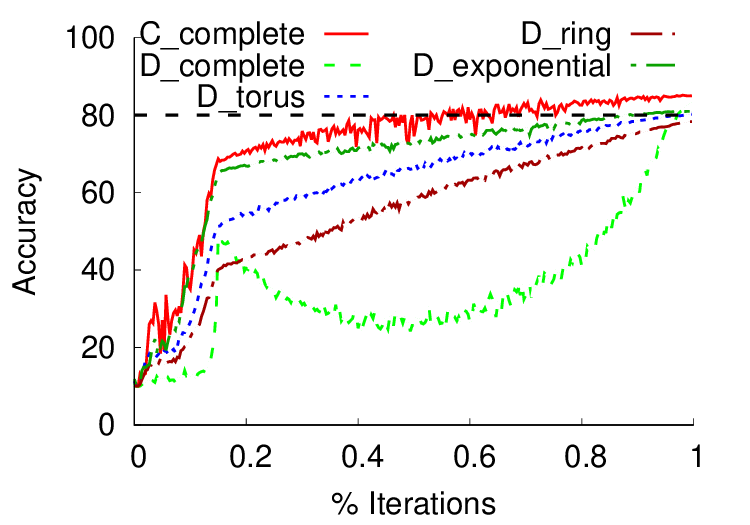}} 
    \subfigure[DensetNet100, 96GPUs\label{fig:a}]{\includegraphics[width=0.245\textwidth]{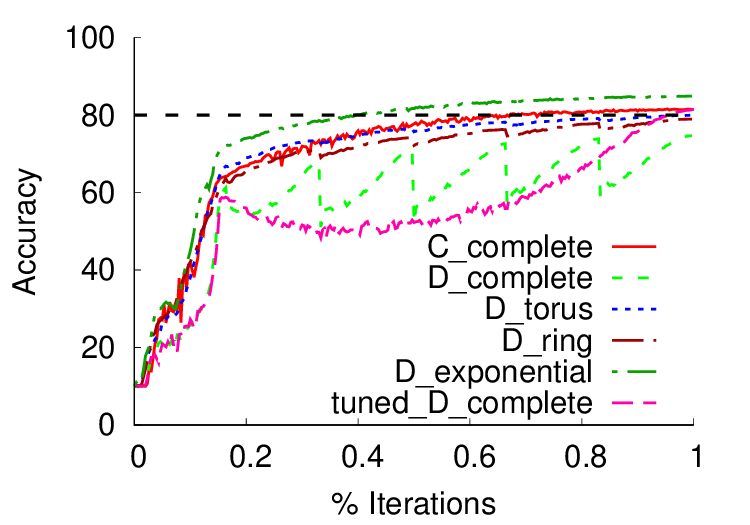}} 
    \subfigure[ResNet50, 96GPUs]{\includegraphics[width=0.245\textwidth]{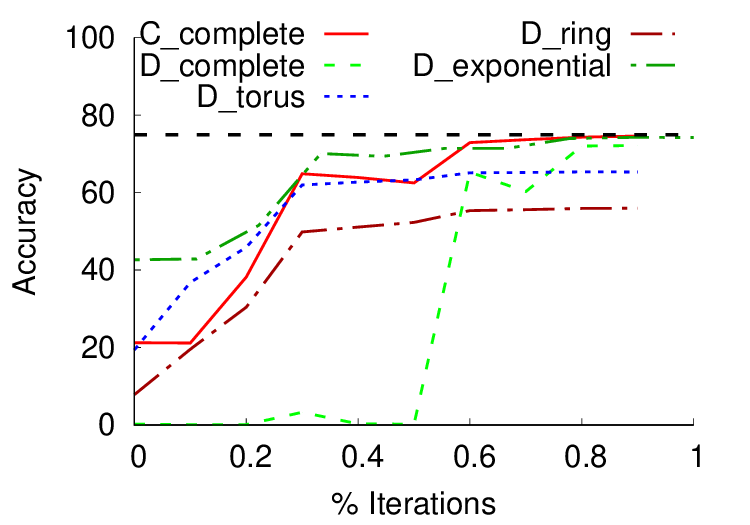}} 
    \subfigure[LSTM, 96GPUs\label{fig:c}]{\includegraphics[width=0.245\textwidth]{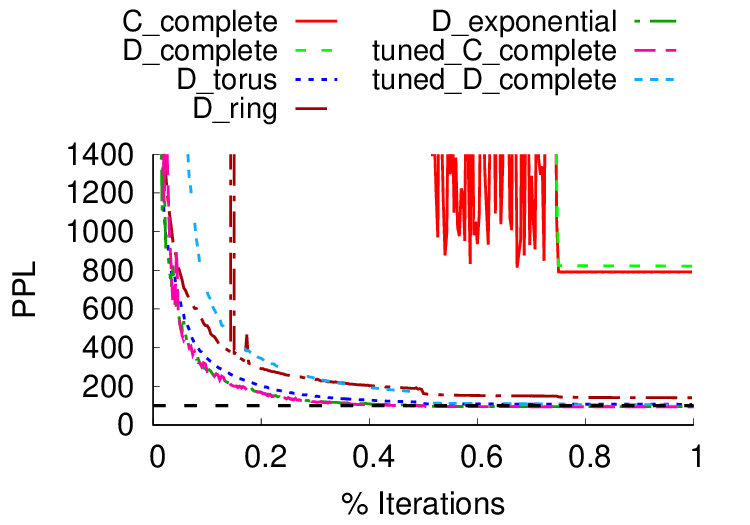}}
    \caption{\figcap{Summary on model accuracy across communication graphs and training scales.} Each subfigure presents the accuracy of the models trained with four SGD implementations (defined in \S\ref{subsec:methodology}) on a fixed number of GPUs for an application.   \iffalse The implementations of \texttt{tuned\_C\_complete} and \texttt{tuned\_D\_complete} are defined in Section~\ref{subsec:accuracy}.\fi}
    \label{fig:test_accuracy}
\end{figure*}

This section discusses model accuracy of centralized and decentralized DL training across communication graphs, training scales, and applications.
Figure \ref{fig:t_accuracy} presents model accuracy of test data (test accuracy) for ResNet50 with two SGD implementations (discussed in \S\ref{subsec:learning_definition}) across training scales. 
Similarly, Figure~\ref{fig:test_accuracy} presents the test accuracy trained on 24 GPUs --- 96 GPUs for all of the four sample applications. We leave results of the test accuracy of ResNet50 with decentralized torus, decentralized exponential, and the results of the test accuracy for the models trained on 12 GPUs to the Appendix (Section~\ref{sec:more}), due to similarity, redundancy and space limitation. 

We first look into the results presented in Figure~\ref{fig:t_accuracy}. It shows clearly that, for a given SGD implementation, when scaling up the training from 12 GPUs to 96 GPUs, the model accuracy decreases in 2\% --- 23.4\% (\texttt{D\_ring}) and in 1.4\% --- 5\% (\texttt{D\_complete}), respectively. We observe a similar trend on all of the decentralized SGD implementations for all of the applications. Thus, we conclude that, similar to centralized learning, decentralized learning also presents the scalability and generality issue: when increasing training scale, model accuracy of decentralized learning decreases no matter which communication graph is in use.

Next, we take a look on the results shown in Figure~\ref{fig:test_accuracy}. 
We first confirm that, when a training scale of a decentralized SGD run increases, the final model accuracy of the run decreases when trained with all of the communication graphs for all of the applications. In particular, when adding the number of GPUs from 12 to 96, the final model accuracy drops 1.2\% --- 17\% for ResNet20, 6.7\% --- 26.5\% for DenseNet100, 1.4\% --- 24.2\% for ResNet50, and 52 --- 462 PPL \footnote{\textcolor{black}{We use PPL to measure training quality of the LSTM experiments. Here, PPL, or named as Perplexity, is a common metric that measures quality of a language model. A lower PPL value represents a better model accuracy.}} for LSTM, respectively.  Again, it confirms the scalability and generality issue in decentralized learning. 

Moreover,  it is clear that, across training scales (all four scales) and applications (all four applications), 81.25\% (13 out of 4$\times$4) of the results show the same pattern on model accuracy with two major facets:
\begin{enumerate*}
\item Centralized learning (\texttt{C\_complete}) consistently delivers the best final accuracy or comparably to the best if not. 
\item For decentralized learning, when there are more connections in a communication graph, the better final accuracy is achieved by the corresponding SGD implementation. In particular, in such a subfigure, the worst to the best convergence rates are delivered by decentralized ring (\texttt{D\_ring}), decentralized torus (\texttt{D\_torus}), decentralized exponential (\texttt{D\_exponential}), and decentralized complete (\texttt{D\_complete}), respectively.
\end{enumerate*}

As is shown clearly in  Figure~\ref{fig:top} and Table~\ref{tab:graph}, exponential graph is very different from the other graphs used in decentralized DL training in many aspects. Nevertheless, all of the graphs follow the same convergence pattern on the number of connections in a communication graph: 
the best, the middle, and the worst convergence rates are delivered by the complete graph (the most connected), the exponential graph and torus (the medium connected), and the rings (the least connected), respectively.    
To summarize, the model accuracy of decentralized DL training is positively correlated to the number of connections in the associated communication graph. The more connections a communication graph has, the better accuracy the decentralized DL training run can finally achieve.  

Beyond the model accuracy obtained at the end, we also notice that, in all of the 16 results for the test accuracy, \texttt{D\_ring} delivers the lowest accuracy consistently when the training progresses at the early stages for DenseNet100, or at the middle stages for ResNet50 and LSTM, or sometimes even at the late stages for ResNet20. In consideration of the good performance at the end \texttt{D\_ring} achieves in 81.25\% of the results, we suspect that, for \texttt{D\_ring}, the poor intermediate performance on model accuracy  might be associated with the conventional configurations on learning rate (Table~\ref{tab:exp}), implying that at the early stages of large-scale decentralized DL training the learning-rate configurations are too large to learn efficiently.

We look into the remaining 8.75\% of the results that do not follow the aforementioned pattern, including DenseNet100 trained on 96 GPUs and LSTM trained on 48 GPUs and 96 GPUs.
Surprisingly, we observe that, \texttt{D\_complete} delivers the lowest convergence rate in the DenseNet100 trained on 96 GPUs. Even more surprisingly, we notice that, for the LSTM trained on 48 and 96 GPUs, neither \texttt{C\_complete} nor \texttt{D\_complete} can converge at the end. Relative to the intermediate poor performance of \texttt{D\_ring} and to the fact that the unconvergence of DenseNet100 and LSTM occurs on relatively large scales (48 and 96 GPUs) , we suspect that the unconvergence might also be related to the conventional linear scaling strategies (discussed in \S\ref{subsec:learning_definition} and presented in Table~\ref{tab:exp}) used for learning rate configurations as they are too large to learn efficiently, and may finally lead to model unconvergence at large scales.   

To verify our conjectures, we fine-tuned the learning rates on \texttt{D\_complete} for DenseNet100 on 96 GPUs, and \texttt{C\_complete} and \texttt{D\_complete} for LSTM on 48 and 96 GPUs, and report the results as \texttt{tuned\_D\_complete} in Figure \ref{fig:a}, and \texttt{tuned\_C\_complete} and \texttt{tuned\_D\_complete} in Figures \ref{fig:b} and \ref{fig:c}, respectively. In the fine-tuned SGDs, instead of using linear scaling, we use the square root scaling to the base learning rate. As is shown clearly, the square root scaling reduced the resulted learning rate significantly. This is a practice adopted by large-scale centralized training as well. The difference is that, for decentralized learning, it seems to be efficient when using the square root scaling at a relatively smaller scale. 

\vspace{2mm}
\noindent {\bf Observation \circled{1}}. Similar to centralized learning, decentralized learning presents the scalability and generality issues when the training scales up. 

\vspace{2mm}
\noindent {\bf Observation \circled{2}}. For decentralized learning, the model accuracy is positively correlated to the number of connections in communication graphs. The more connections a communication graph has, the better accuracy the corresponding decentralized training run can achieve. 

\vspace{2mm}
\noindent {\bf Observation \circled{3}}. For decentralized learning, the policies on learning rate configurations might be different from centralized learning. More specifically, compared to centralized learning, the square root scaling can start improving the model accuracy at a relatively smaller scale.  

\subsection{Analysis on Variances of Parameter Tensors}
\label{subsec:variance}

\begin{figure*}[t]
    \centering
    
    \subfigure[ResNet20, \texttt{D\_ring}]{\includegraphics[width=0.235\textwidth]{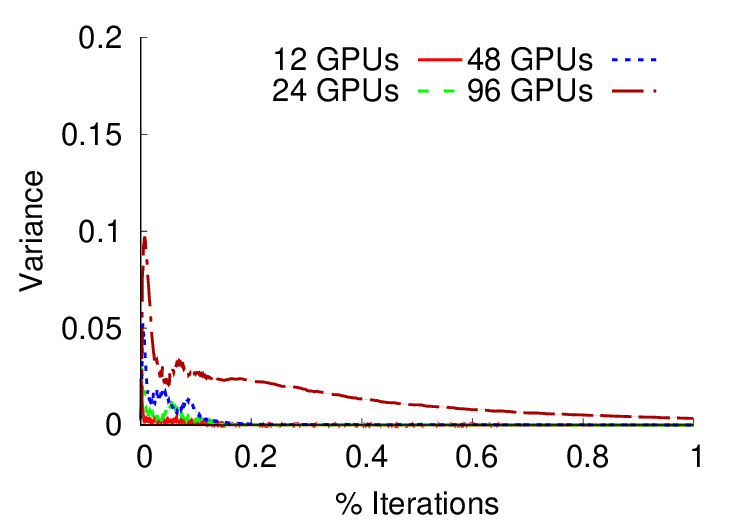}} 
    \subfigure[DenseNet100, \texttt{D\_torus}]{\includegraphics[width=0.235\textwidth]{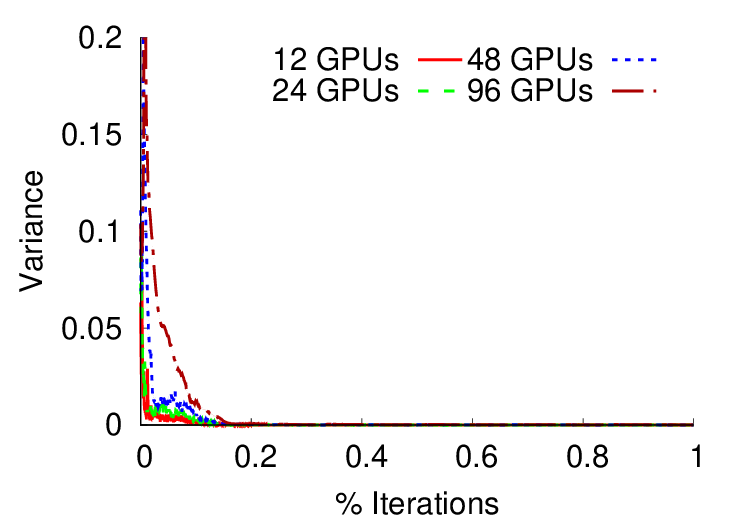}} 
    \subfigure[ResNet50,  \texttt{D\_complete}]{\includegraphics[width=0.235\textwidth]{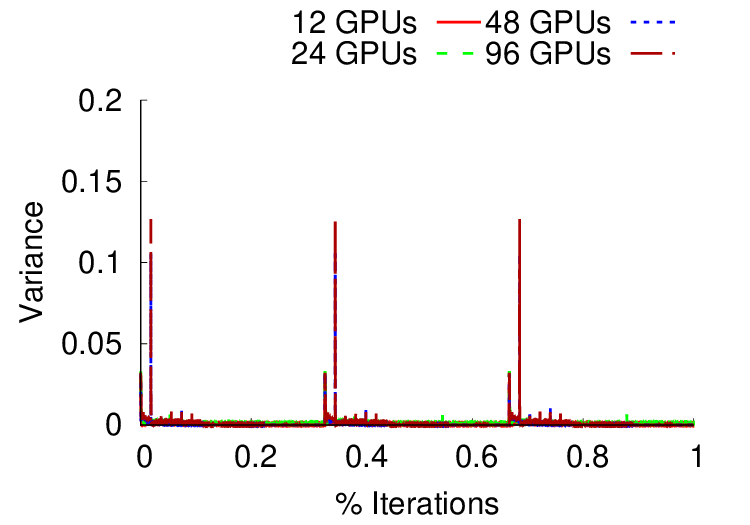}} 
    \subfigure[LSTM,  \texttt{D\_ring}]{\includegraphics[width=0.235\textwidth]{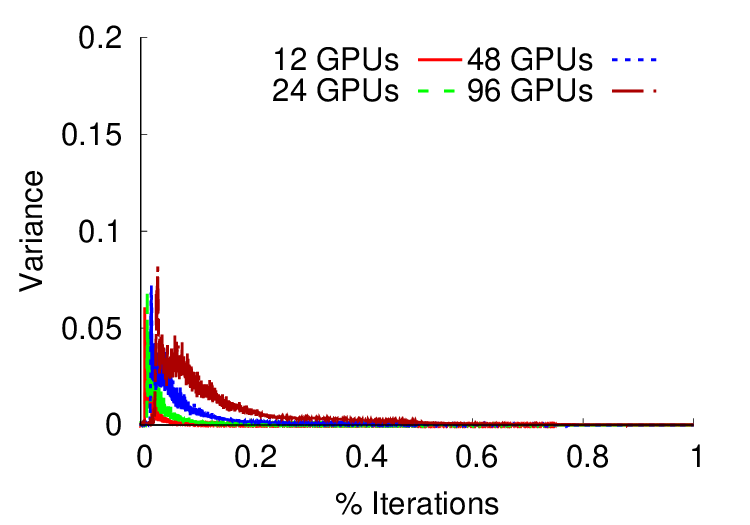}} 

    \subfigure[ResNet20, 96 GPUs]{\includegraphics[width=0.235\textwidth]{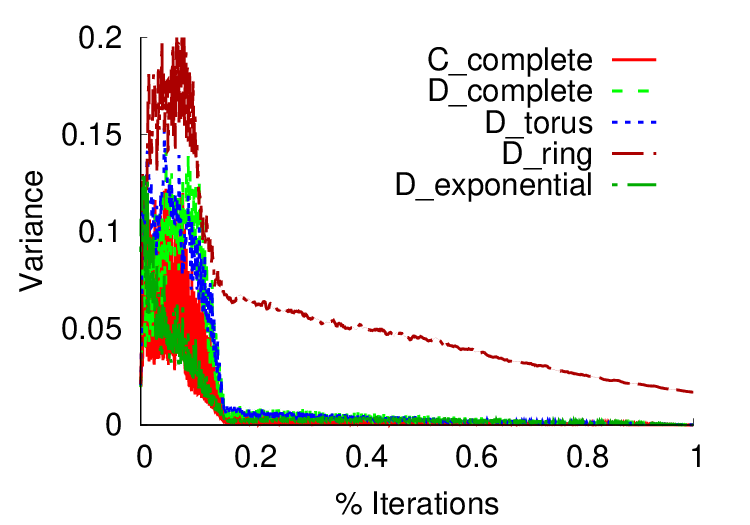}} 
    \subfigure[DenseNet100, 96 GPUs\label{fig:e}]{\includegraphics[width=0.235\textwidth]{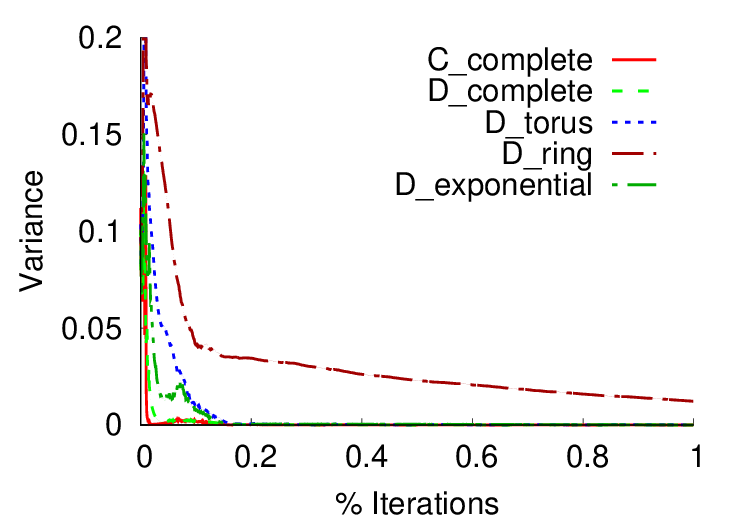}} 
    \subfigure[ResNet50, 96 GPUs\label{fig:d}]{\includegraphics[width=0.235\textwidth]{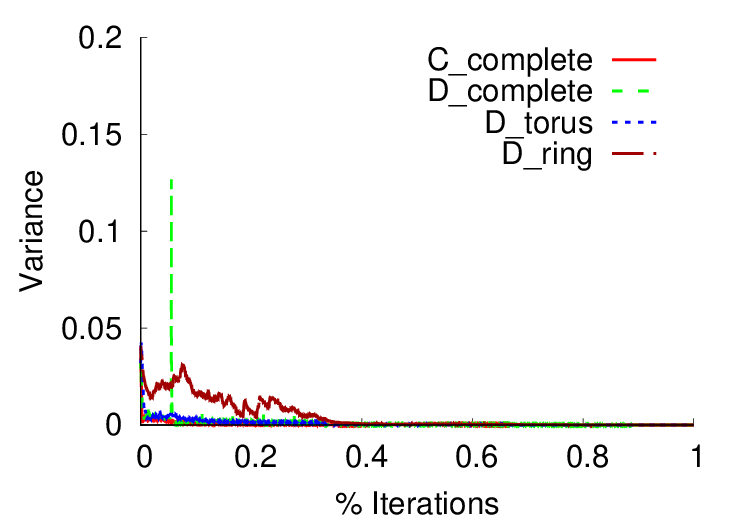}} 
    \subfigure[LSTM, 96 GPUs]{\includegraphics[width=0.235\textwidth]{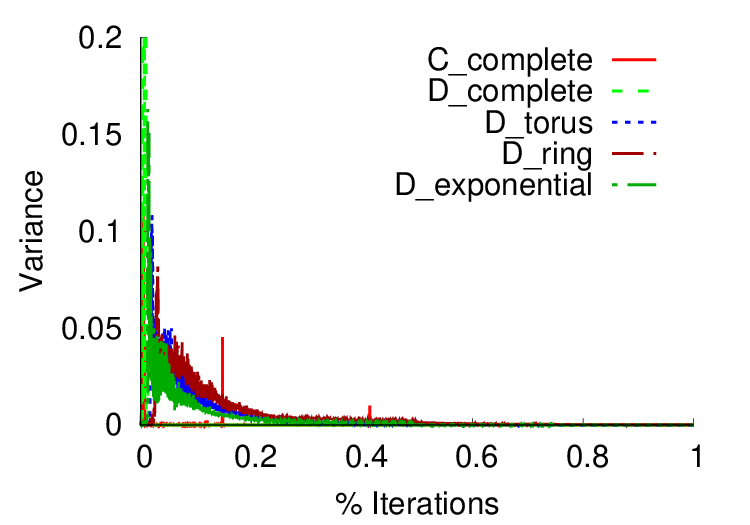}}   
      
    \caption{\figcap{Gini coefficients of single parameters across communication graphs and training scales.}}
    \label{fig:variance_sample}
\end{figure*}

\begin{figure*}[t]
    \centering
    
    \subfigure[ResNet20, 24 GPUs]{\includegraphics[width=0.245\textwidth]{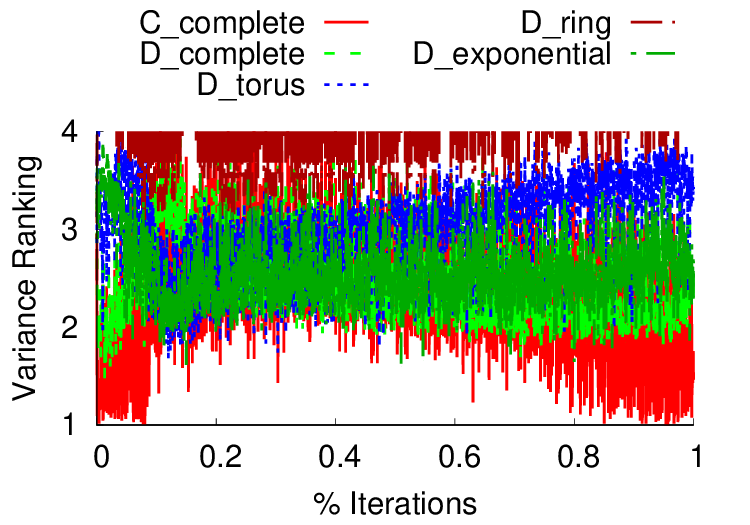}} 
     \subfigure[DenseNet100, 24 GPUs]{\includegraphics[width=0.245\textwidth]{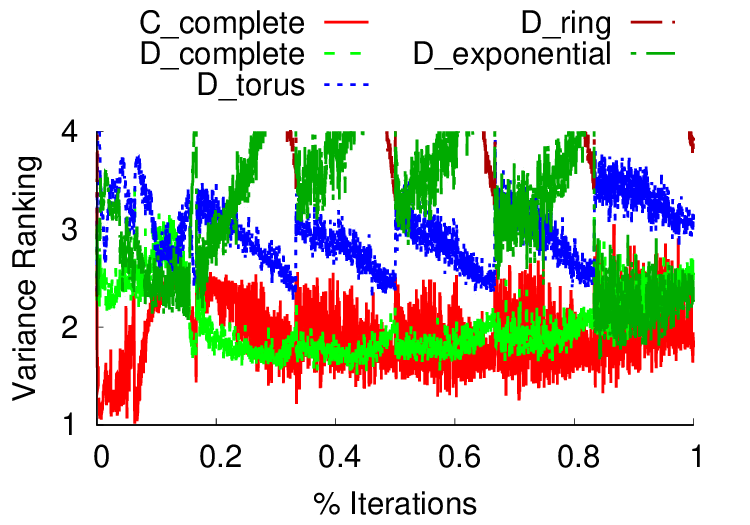}} 
    \subfigure[ResNet50, 24 GPUs]{\includegraphics[width=0.245\textwidth]{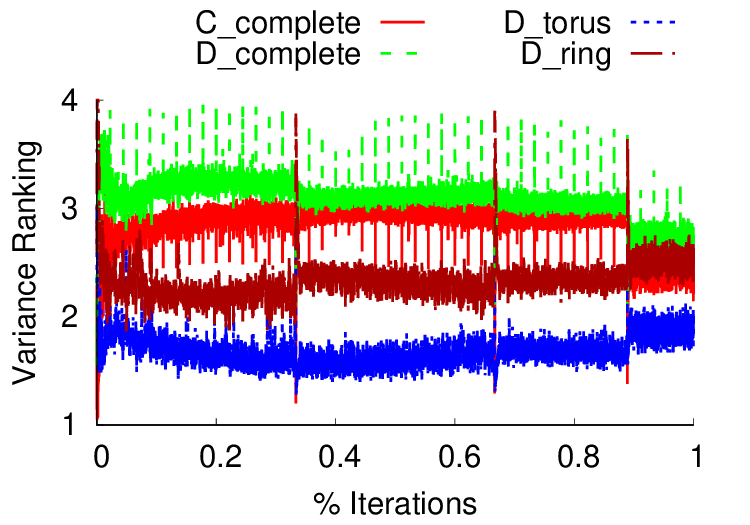}} 
    \subfigure[LSTM, 24 GPUs]{\includegraphics[width=0.245\textwidth]{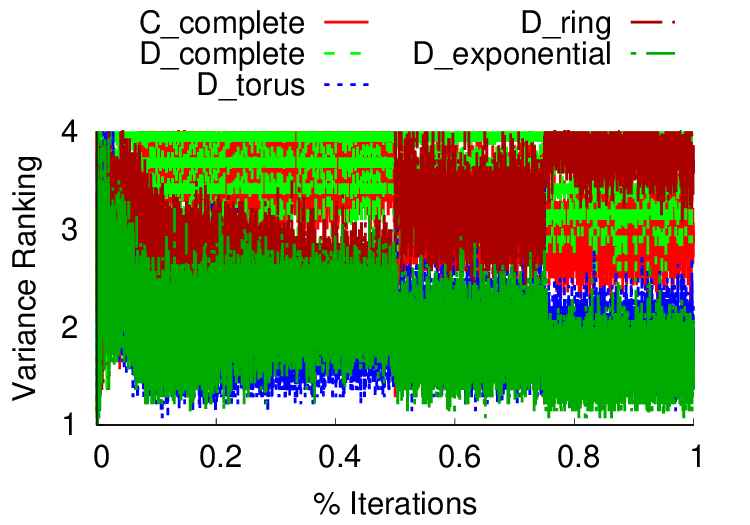}}

    \subfigure[ResNet20, 48 GPUs]{\includegraphics[width=0.245\textwidth]{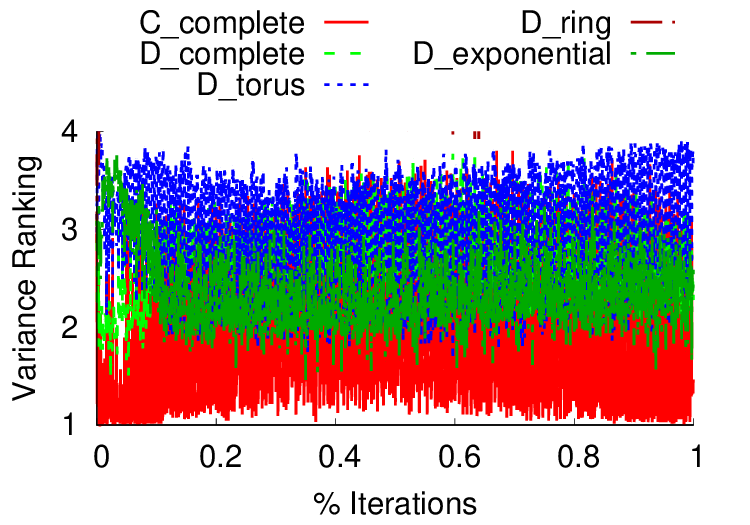}} 
     \subfigure[DenseNet100, 48 GPUs]{\includegraphics[width=0.245\textwidth]{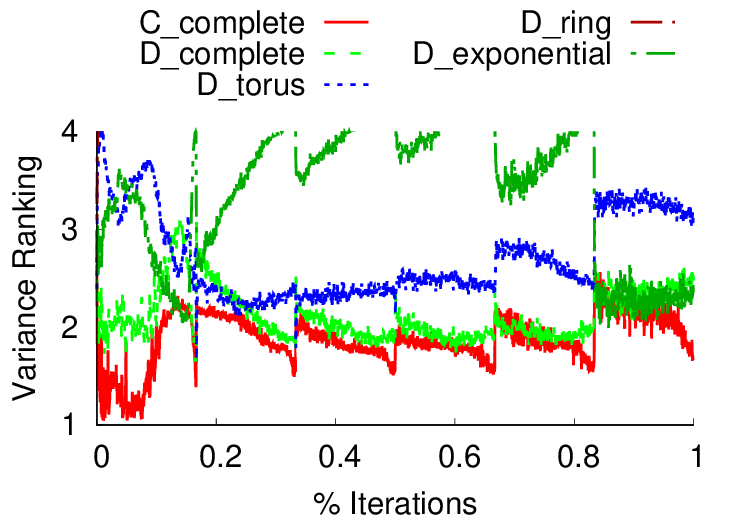}} 
     \subfigure[ResNet50, 48 GPUs]{\includegraphics[width=0.245\textwidth]{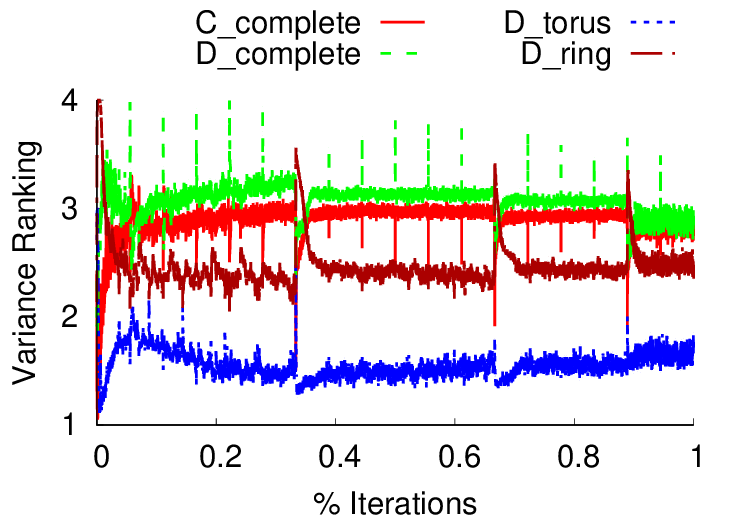}} 
     \subfigure[LSTM, 48 GPUs]{\includegraphics[width=0.245\textwidth]{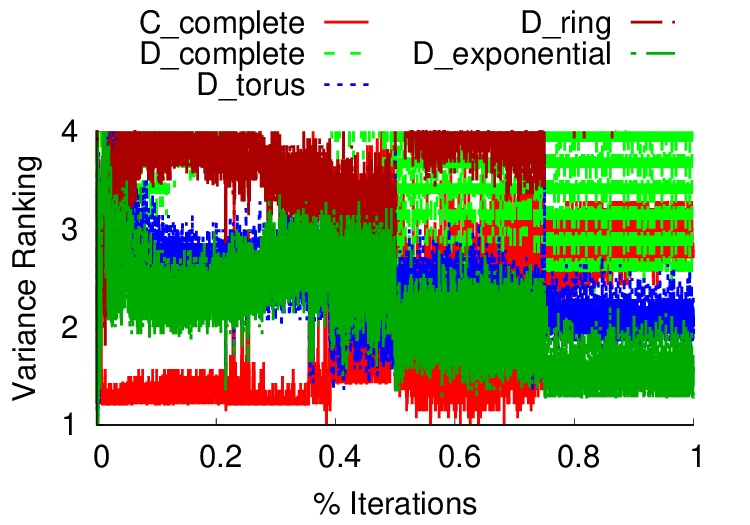}}

     \subfigure[ResNet20, 96 GPUs]{\includegraphics[width=0.245\textwidth]{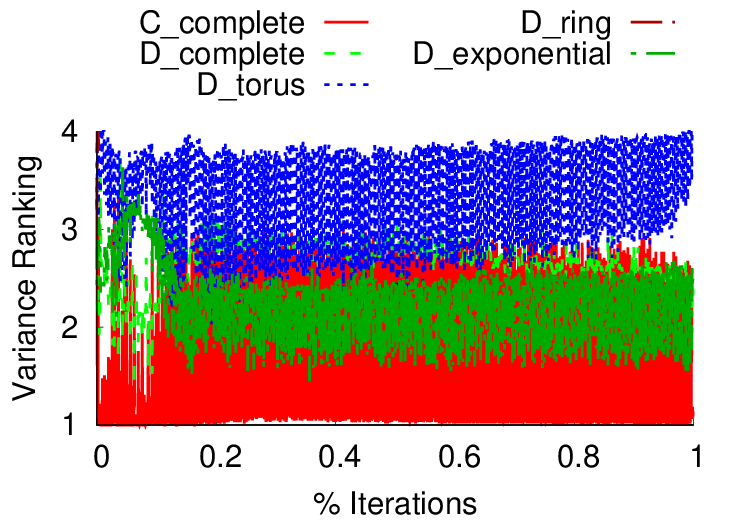}} 
     \subfigure[DenseNet100, 96 GPUs]{\includegraphics[width=0.245\textwidth]{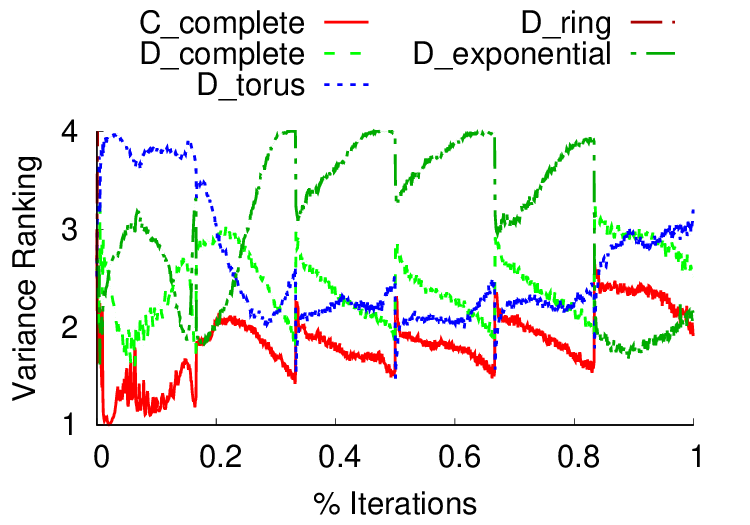}} 
     \subfigure[ResNet50, 96 GPUs]{\includegraphics[width=0.245\textwidth]{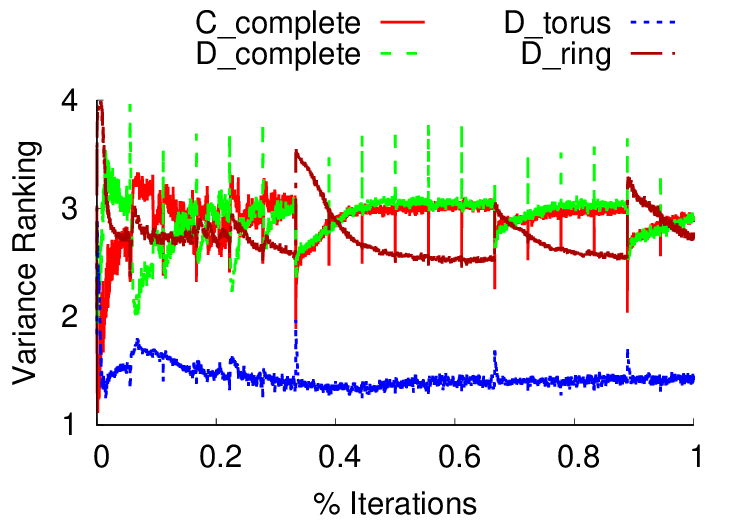}} 
      \subfigure[LSTM, 96 GPUs]{\includegraphics[width=0.245\textwidth]{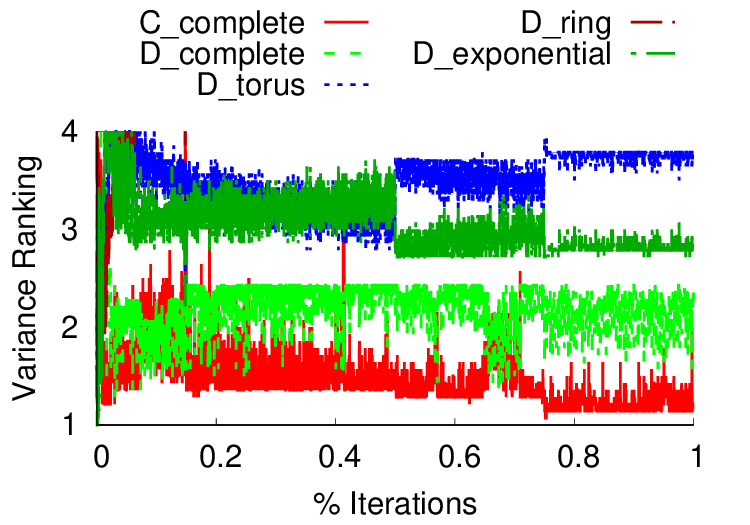}}
    \caption{\figcap{Summary of variance ranks of parameter tensors.} We define the variance ranks of parameter tensors in \S\ref{subsec:variance}.}
    \label{fig:parameter_ranking}
\end{figure*}

This section discusses the variances of parameter tensors across GPUs for different SGD implementations, with different communication graphs, and on different scales. 
 We are motivated by the previous observations \cite{ddp} that, compared to the gradient tensor averaging in centralized learning, decentralized learning exhibits vast differences on the training results across GPUs. These vast differences cause conflict directions on gradient descent throughout iterations and may lead to model unconvergence eventually.

To conduct an in-depth analysis on the differences of local training results and motivated by the recent studies on understanding the DNN training internals using gradient norms \cite{ross2017neural, ross2018improving, chen2018gradnorm, zhang2019gradient}. We analyze the variances of parameter tensors across model replicas before the parameter-tensor is averaged in an iteration and executed by following different SGD implementations. 
To ensure the unbiased evaluations about variances of parameter tensors, we investigate the variances for all of the parameters in all of the experiments for all of the applications with four widely used metrics, including gini coefficient, index of dispersion, coefficient of variation, and quartile coefficient of dispersion. 

We observe that, the results of different metrics present the same trends and patterns consistently. Thus, we report the results with gini coefficient and leave the results of the other metrics to our anonymous repository\footref{foot}. We skip the report of variances of parameter tensors on \texttt{D\_exponential} for ResNet50 since we see exponential graphs present the similar/same results in both model accuracy (Section~\ref{subsec:accuracy}) and in the variance analysis across applications on one of the popular metrics. We iterate the rest in the Appendix.

\vspace{2mm}
\noindent {\bf Gini coefficient}, originated from economics, is defined with the mean $\mu$ and the probability distribution function $p(x)$: $$\frac{1}{2\mu}\int^{\infty}_{-\infty}\int^{\infty}_{-\infty}p(x)p(y)|x-y|~dxdy$$
We present the ranges of the parameter-tensor dimensions in Table~\ref{tab:bench}  in the column \texttt{Tensor Size}.

We start from understanding the variances of single parameters.  Figure~\ref{fig:variance_sample} presents the results of the variances of a sample parameter for each application. In particular, for each application, we report the results of an SGD implementation across training scales and the results of the training runs on 96 GPUs with all of the 5 SGD implementations. 
We observe similar trends on the other not reported results.

Relative to the results in Figure~\ref{fig:test_accuracy}, Figure~\ref{fig:variance_sample} first shows that, across iterations and SGD implementations, when the high variances show up in a parameter tensor, the corresponding model accuracy is low.  Moreover, we see that, for different parameters across iterations in the same training run, the parameter-tensor variances show similar patterns on low and high values, indicating that an SGD implementation affects the training efficiency of the parameters in the run similarly. In summary, we conclude that, for decentralized learning, the variances of parameter tensors are highly likely related to model accuracy: the higher variances the parameter tensors present, the lower model accuracy the corresponding decentralized training run delivers. 

Moreover, Figure~\ref{fig:variance_sample} also shows that, across SGD implementations and communication graphs, when a training run progresses, the variances of parameter tensors decrease. This observation is intuitive as the model converges progressively. However, it is worth noticing that, relative to the test accuracy given in Figure~\ref{fig:c}, although  delivering equally low variances in Figures~\ref{fig:d}, \texttt{D\_ring} and \texttt{D\_torus} obtain relatively low resulted convergence rate when trained on 96 GPUs for ResNet50. We also observe the same trend on the other low model accuracy runs, such like \texttt{D\_complete} on the 48 and 96 GPUs training runs for LSTM. This observation seemingly suggests that a decentralized DL training run obtains the local optimal at the end, but at the same time delivers relatively low variances of individual parameter tensors. Relative to the aforementioned observation on the correlation between model accuracy and parameter-tensor variances, we conclude that, in decentralized learning, the degree of parameter-tensor variances is a sufficient but not necessary condition for model accuracy: the high variances suggests the poor model accuracy, but the low variances cannot guarantee the high model accuracy at the end.       

Beyond the observation on the correlation between model accuracy and parameter-tensor variances derived from Figure~~\ref{fig:variance_sample}, we further find  that, for an application at a training scale, 
the variances of parameter tensors associated with different communication graphs present vast differences at the start of the training as \texttt{D\_ring} consistently presents the highest variances of parameter tensors, and \texttt{C\_complete} and/or \texttt{D\_complete} consistently presents the lowest. 
But when the training progresses, these variance differences across communication graphs diminish. Relative to the accuracy report presented in Figure~\ref{fig:test_accuracy},  it suggests that, at the beginning of a training run, the training benefits more from a graph where each node is connected to a higher number of neighbors. But when the training progresses, this benefit diminishes. This observation indicates that, for decentralized DL training, an adaptive approach might be more beneficial: at the start the training benefits from a highly connected graph for high model accuracy. When it progresses, the training benefits more from the graphs with fewer connections that offer lower communication costs with negligibly less or even no loss of accuracy.    

\vspace{2mm}
\noindent {\bf Ranking Analysis}. To further investigate the correlation between model accuracy and parameter-tensor variance, we integrate the performance of variances of all parameter tensors by assigning them ranks. Built on top of ranking analysis, we can compare the 5 SGD implementations when they are executed at the same iteration, on the same training scale, and for the same application.

For each SGD implementation, in each iteration, we assign the implementation a ranking value in 1 --- 4 based on the value of its gini coefficient. The ranking values of 1, 2, 3, 4 represent the implementation delivers the lowest, the medium low, the medium high, and the highest variance, respectively. Although these variance ranks filter out the value differences among the variances, it makes the variances across parameters comparable and integrable.  

Figure~\ref{fig:parameter_ranking} presents the results. Relative to Figure~\ref{fig:test_accuracy}, it shows clearly that, for the ranking summary of ResNet20, the ranking order is consistent to the model accuracy: across iterations and communication graphs, \texttt{C\_complete},  \texttt{D\_complete}, \texttt{C\_exponential}, \texttt{C\_torus}, \texttt{C\_ring} are ranked in an ascending order, and their model accuracies are in a descending order. Different from the pattern shown in ResNet20, the other three applications present different correlation between variance ranks and model accuracy. These results, again, confirm that, the degree of parameter-tensor variances is correlated to model accuracy.

\vspace{2mm}
\noindent {\bf Observation \circled{4}}. The degree of parameter-tensor variance is correlated to the number of connections of a communication graph and to the model accuracy, especially at the early stage of a training run. In particular, when such a run starts, the more connections a communication graph has, the lower variances parameter tensors present, and at the same time a higher model accuracy the run obtains. When the run progresses, these cross-graph variance differences diminish with no loss of model accuracy, suggesting the benefit of the less connected graphs at a later stage of the training as they offer low communication cost with no sacrifice on accuracy.


\vspace{2mm}
\noindent {\bf Observation \circled{5}}. For decentralized DL training, an adaptive approach might be more beneficial than a static solution with a fixed communication graph. In particular, at the start of a training run, it benefits from a highly connected graph for high model accuracy. When the training progresses, it benefits more from a graph with fewer connections that offer lower communication cost with no loss of accuracy.

\section{Adaptive Decentralized Data Parallelism Training}
\label{sec:solution}

This section introduces {\it \name}, an adaptive approach of decentralized data parallel training. 
Built upon decentralized SGD methods, \name is motivated by our observations derived from a group of controlled benchmarking experiments (\S\ref{sec:analysis}) and is proposed to achieve high model accuracy and low communication cost for large scale DNN training in production use.

Different from the existing decentralized DL training approaches that use fixed communication graphs during the training~\cite{lian2017can, koloskova2019decentralized,koloskova2020decentralized}, \name is an adaptive decentralized SGD process. For a DNN training run managed by Ada, the number of connections per node in the communication graph in use is reduced across iterations to obtain the high accuracy achieved by the highly connected graphs at the early stage of the training and obtain the low communication cost achieved by the lowly connected graphs later on with no loss of model accuracy (discussed in \S\ref{subsec:variance}). 

 To the best of our knowledge, \name is the first effort of adaptive decentralized DL training and the first solution in decentralized learning that achieves equally or comparably good accuracy as centralized learning
does for all sample applications, even when training ResNet50 for ImageNet-1K on the scale of 1008 GPUs (\S\ref{subsec:validation}).

\begin{figure*}[t!]
\centering
\subfigure[$k = 4$]{
\begin{tikzpicture}[scale=0.7]
  \path (-90+0*40:2) node (n0) [neighbor node] {};
  \path (-90+1*40:2) node (n1) [self node] {};
  \path (-90+2*40:2) node (n2) [neighbor node] {};
  \path (-90+3*40:2) node (n3) [neighbor node] {};
  \path (-90+4*40:2) node (n4) [neighbor node] {};
  \path (-90+5*40:2) node (n5) [neighbor node] {};
  \path (-90+6*40:2) node (n6) [neighbor node] {};
  \path (-90+7*40:2) node (n7) [neighbor node] {};
  \path (-90+8*40:2) node (n8) [neighbor node] {};
  
  \foreach \i in {0,...,8}
    \foreach \j in {0,...,8}
      \path [net connect]
        (n\i) -- (n\j);
        
  \foreach \i in {0,...,8}
      \path [net thick connect]
        (n\i) -- (n1);
\end{tikzpicture}
}
\subfigure[$k = 3$]{
\begin{tikzpicture}[scale=0.7]
  \path (-90+0*40:2) node (n0) [neighbor node] {};
  \path (-90+1*40:2) node (n1) [self node] {};
  \path (-90+2*40:2) node (n2) [neighbor node] {};
  \path (-90+3*40:2) node (n3) [neighbor node] {};
  \path (-90+4*40:2) node (n4) [neighbor node] {};
  \path (-90+5*40:2) node (n5) [net node] {};
  \path (-90+6*40:2) node (n6) [net node] {};
  \path (-90+7*40:2) node (n7) [neighbor node] {};
  \path (-90+8*40:2) node (n8) [neighbor node] {};
  
  \path [net thick connect] (n0) -- (n1) -- (n2);
  \path [net connect] (n2) -- (n3) -- (n4) -- (n5) -- (n6) -- (n7) -- (n8) -- (n0);
  
  \path [net connect] (n0) -- (n2);
  \path [net connect] (n2) -- (n4);
  \path [net connect] (n3) -- (n5);
  \path [net connect] (n4) -- (n6);
  \path [net connect] (n5) -- (n7);
  \path [net connect] (n6) -- (n8);
  \path [net connect] (n7) -- (n0);
  
  \path [net connect] (n0) -- (n3);
  \path [net connect] (n2) -- (n5);
  \path [net connect] (n3) -- (n6);
  \path [net connect] (n4) -- (n7);
  \path [net connect] (n5) -- (n8);
  \path [net connect] (n6) -- (n0);

  \path [net connect] (n8) -- (n2);
  \path [net connect] (n2) -- (n5);
  \path [net connect] (n3) -- (n6);
  \path [net connect] (n4) -- (n7);
  \path [net connect] (n5) -- (n8);
  \path [net connect] (n6) -- (n0);
  
  \path [net thick connect] (n1) -- (n3);
  \path [net thick connect] (n8) -- (n1);
  \path [net thick connect] (n1) -- (n4);
  \path [net thick connect] (n7) -- (n1);
\end{tikzpicture}
}
\subfigure[$k = 2$]{
\begin{tikzpicture}[scale=0.7]
  \path (-90+0*40:2) node (n0) [neighbor node] {};
  \path (-90+1*40:2) node (n1) [self node] {};
  \path (-90+2*40:2) node (n2) [neighbor node] {};
  \path (-90+3*40:2) node (n3) [neighbor node] {};
  \path (-90+4*40:2) node (n4) [net node] {};
  \path (-90+5*40:2) node (n5) [net node] {};
  \path (-90+6*40:2) node (n6) [net node] {};
  \path (-90+7*40:2) node (n7) [net node] {};
  \path (-90+8*40:2) node (n8) [neighbor node] {};
  
  \path [net thick connect] (n0) -- (n1) -- (n2);
  \path [net connect] (n2) -- (n3) -- (n4) -- (n5) -- (n6) -- (n7) -- (n8) -- (n0);
  
  \path [net connect] (n0) -- (n2);
  \path [net connect] (n2) -- (n4);
  \path [net connect] (n3) -- (n5);
  \path [net connect] (n4) -- (n6);
  \path [net connect] (n5) -- (n7);
  \path [net connect] (n6) -- (n8);
  \path [net connect] (n7) -- (n0);
  
  \path [net thick connect] (n1) -- (n3);
  \path [net thick connect] (n8) -- (n1);
\end{tikzpicture}
}
\subfigure[$k = 1$]{
\begin{tikzpicture}[scale=0.7]
  \path (-90+0*40:2) node (n0) [neighbor node] {};
  \path (-90+1*40:2) node (n1) [self node] {};
  \path (-90+2*40:2) node (n2) [neighbor node] {};
  \path (-90+3*40:2) node (n3) [net node] {};
  \path (-90+4*40:2) node (n4) [net node] {};
  \path (-90+5*40:2) node (n5) [net node] {};
  \path (-90+6*40:2) node (n6) [net node] {};
  \path (-90+7*40:2) node (n7) [net node] {};
  \path (-90+8*40:2) node (n8) [net node] {};
  
  \path [net thick connect] (n0) -- (n1) -- (n2);
  \path [net connect] (n2) -- (n3) -- (n4) -- (n5) -- (n6) -- (n7) -- (n8) -- (n0);
\end{tikzpicture}
}
\caption{\figcap{A simple example of a ring-lattice evolution in Ada.} Here, the ring lattice consists of 9 nodes and the coordination number $k$ is varied from 4 to 1. $k$ is defined in Section~\ref{sec:solution:design_impl}.  Clearly, in the corresponding decentralized DL training process, the communication graph varies from a fully connected complete graph to a sparse ring. }. 
\label{fig:ring_lattice}
\end{figure*}
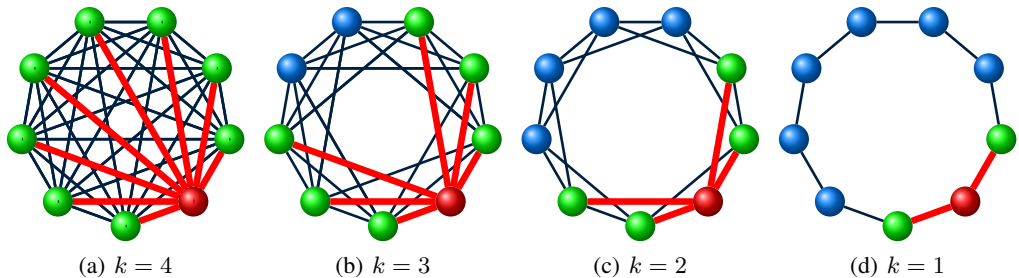

\subsection{Design and Implementation}
\label{sec:solution:design_impl}

The key idea behind \name is to adapt the number of connections of a communication graph during a decentralized SGD process. To realize a varying communication graph easily and efficiently, we choose ring lattice as the base to structure the varying graph. In particular, ring lattice is first introduced to build a random graph with good properties for  social network analysis. More recently, it has been widely used in the domain of AI/ML for its simplicity, stability and flexibility~\cite{ringlatt1,ringlatt2}. 

\vspace{2mm}
\noindent {\bf Ring Lattice}. A ring lattice topology is evolved from a ring topology. Different from a ring in which each node consists of two neighbors with each neighbor one hop away from the node, a ring lattice is built with a {\it coordination number} $k$ in which each node consists of $2k$ neighbors with each neighbor 1 --- $k$ hops away from the node. Clearly, for a ring lattice with $n$ nodes, the number of connections per node can vary significantly when  $k$ varies. 


Motivated by the observations in Section~\ref{subsec:variance}, Ada realizes an adaptive communication graph by starting a training process with a highly connected ring lattice and decreases the value of $k$ associated with the lattice across iterations. Figure~\ref{fig:ring_lattice} presents a simple example of the adaptive approach with 9 nodes in the lattice.

Assuming that each node in the lattice represents a GPU performing a decentralized SGD process locally. At the start, we set $k$ = 4 with each node connecting to the remaining 8 ($2\times4$) nodes and accordingly let the training average parameters based on a complete graph (Figure~\ref{fig:ring_lattice} (a)). When the training progresses, we reduce the value of $k$ from 4 to 1. Eventually, when $k=1$, we make each node connect to 2 ($2\times1$) neighbors and accordingly let the training average parameters based on a ring (Figure~\ref{fig:ring_lattice} (d)).       
\begin{algorithm}[t]
\caption{\figcap{Adapting the ring lattice in Ada with hyperparameters $k$ and $\gamma_k$.}}
\begin{algorithmic}[1]
\For{epoch=1, nepochs}
\State{$k$ $\gets$ max($k_0$-int($\gamma_k$\,epoch), 2)}
\State{graph $\gets$ array(nGPUs, nGPUs)}
\For{$i=0$, nGPUs-1}
\State{graph[$i$][$i$] $\gets$ $1/(k+1)$}
\For{$j=-k//2, k//2$}
\If{$j!=0$}
\State{graph[$i$][$(i+j)$\%nGPUs)]=$1/(k+1)$}
\EndIf
\EndFor
\EndFor
\State{decentralized\_training(epoch, graph)}
\EndFor
\end{algorithmic}\label{algo:ada}
\end{algorithm}
Compared to the centralized DL training and the decentralized SGD with fixed communication graphs, our decentralized adaptive approach Ada introduces two new hyperparameters in the training process, managing the coordination numbers ($k$) of ring lattice and the corresponding decay rate ($\gamma_k$) of $k$, respectively.

Algorithm~\ref{algo:ada} denotes the implementation details of how we reduce the number of connections per node in an engaged ring lattice across training iterations by varying $k$ based on $\gamma_k$. In particular, we start from an initial state of a ring lattice with a large $k_0$ for high model accuracy. Next, we employ a linear function:
$k = max(k_0 - \texttt{int}(\gamma_k\, \texttt{Epoch}), 1)$, where $k_0$ represents the initial value of $k$.
This function is introduced to decrease the value of $k$ at each training epoch, and populate the value of $k$ to each GPU for updating the ring lattice topology per GPU locally. We choose a linear function since it is simple, efficient, and widely used in the domain of AI/ML for hyperparameter tuning.    


\begin{figure*}[t!]
  \centering
    \subfigure[ResNet20]{\includegraphics[scale=0.3]{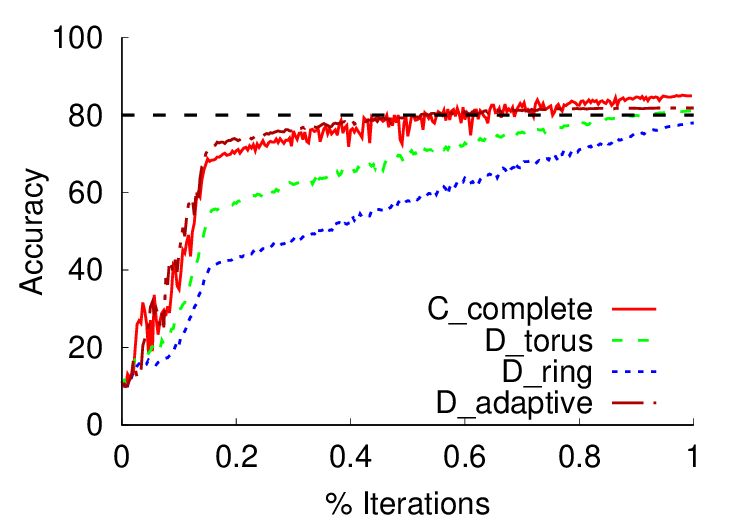}}
    \subfigure[DenseNet100]{\includegraphics[scale=0.3]{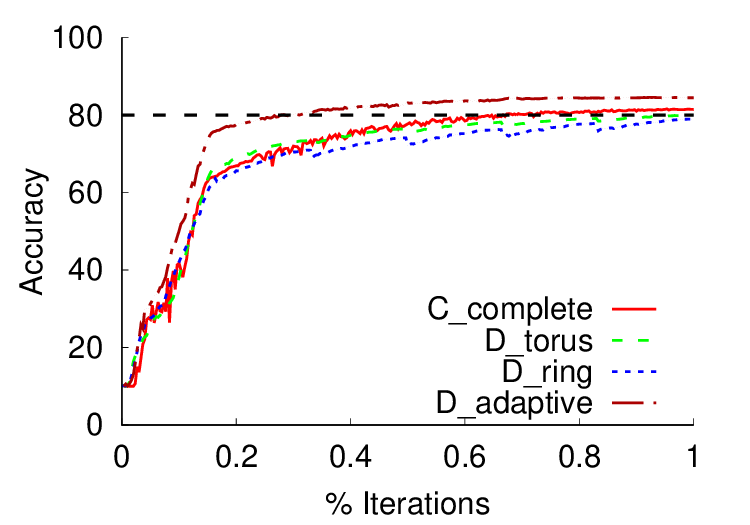}}
    \subfigure[LSTM]{\includegraphics[scale=0.3]{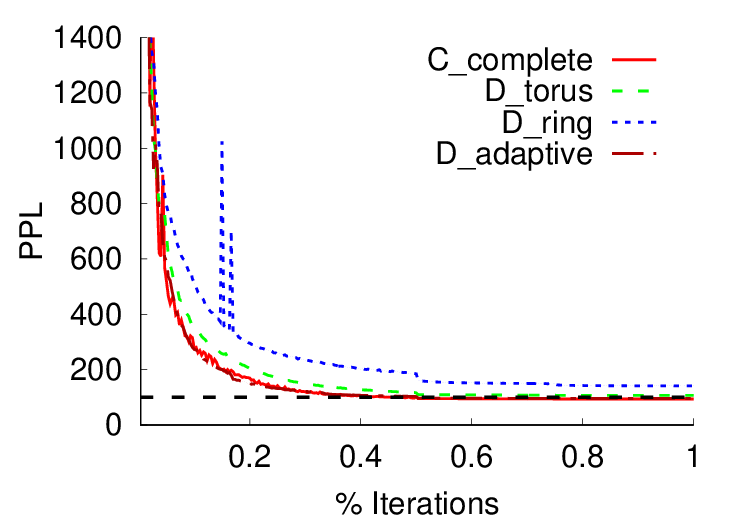}}
    \subfigure[ResNet50]{\includegraphics[scale=0.3]{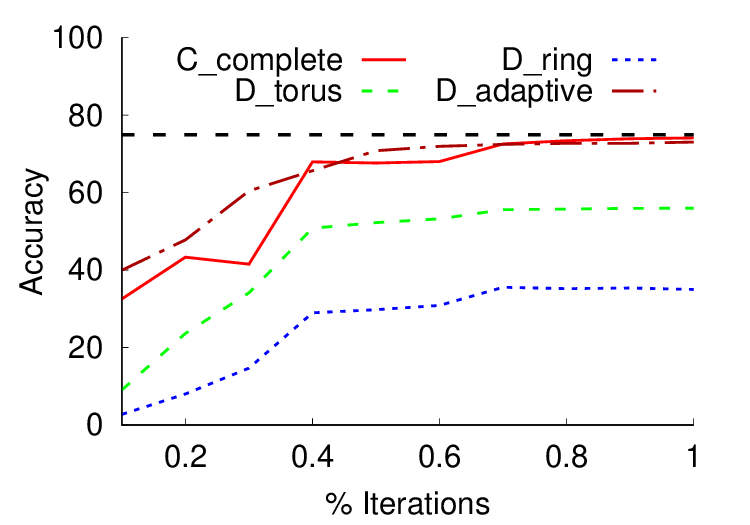}\label{fig:resnet50}}
  \caption{\figcap{Distributed training of ResNet20 and DenseNet100 on CIFAR10, and LSTM on WikiText2, using 96 GPUs; ResNet50 on ImageNet-1k using 1008 GPUs.} 
  Dash line indicates a commonly acceptable test accuracy (higher is better) or perplexity (PPL, lower is better).}
 \label{fig:resnet20}
\end{figure*}

\begin{table}[t!]
\centering
\begin{tabular}{ccccc}
\toprule
     & \begin{tabular}[c]{@{}c@{}}ResNet20\\ 96 GPUs\end{tabular} & \begin{tabular}[c]{@{}c@{}}ResNet50\\ 1008 GPUs\end{tabular} & \begin{tabular}[c]{@{}c@{}}DenseNet100\\ 96 GPUs\end{tabular} & \begin{tabular}[c]{@{}c@{}}LSTM\\ 96 GPUs\end{tabular} \\ \midrule
$k_0$ & 10                                                         & 112                                                        & 10                                                              & 10                                                     \\
$\gamma_k$ & 0.02                                                         & 1                                                        & 0.02                                                              & 0.02                                                     \\\bottomrule
\end{tabular}
\caption{\figcap{Tuning parameters of ring lattice used in our decentralized adaptive approach Ada.} \iffalse Here, $k_0$ is the initial value of $k$. $k$ and $\gamma_k$ are defined in Section~\ref{sec:solution:design_impl}.\fi }\label{tab:ringlatt}
\end{table}

\subsection{Validation}
\label{subsec:validation}
We evaluate our decentralized adaptive approach Ada on both image classification and language modeling tasks on the Summit supercomputer. The experiment settings are similar to those of the white-box analyses listed in Table~\ref{tab:bench} and discussed in Section~\ref{subsec:environment}. 
Different from the system setting used in the benchmarking of centralized and decentralized DL training, we evaluate Ada on the training scale of ResNet50 expanded up to 1008 GPUs. Table~\ref{tab:ringlatt} summarizes the tuning parameters of ring lattice used in our approach. $k_0$ and $\gamma_k$ are defined in Section~\ref{sec:solution:design_impl}. 

We show the performance of our approach in comparison to centralized complete, decentralized ring, and decentralized torus (defined in \S\ref{subsec:learning_definition}). In Figure~\ref{fig:resnet20} (a) and (b), we plot the test accuracy in the distributed training on CIFAR10 using 96 GPUs. For both ResNet20 and DenseNet100 models, our solution (\texttt{D\_adaptive}) shows the fastest convergence to the acceptable accuracy (the horizontal dash line).     
For a larger model such as a 28.95M parameters LSTM \cite{lstm}, both decentralized ring and torus fail to converge (perplexity, i.e. PPL, below 100) at the scale of 96 GPUs, as shown in Figure~\ref{fig:resnet20}(c). Our solution again converges faster with a PPL of 95.7 at the 300 epoch.    
For a larger model such as a 28.95M parameters LSTM \cite{lstm}, both decentralized ring and torus fail to converge (perplexity, i.e. PPL, below 100) at the scale of 96 GPUs, as shown in Figure~\ref{fig:resnet20}(c). Our solution again converges faster with a PPL of 95.7 at the 300 epoch.

The state-of-the-art centralized training of ResNet50 on ImageNet-1k \cite{mlperf} is at a scale of 1K GPUs. In comparison, we show our decentralized solution using 1008 GPUs in Figure~\ref{fig:resnet50}. As far as we know, this is the largest scale that has ever been performed for decentralized training. At this scale, the obtained top1 classification accuracy for ring- and torus-based decentralized SGD are 35\% and 56\%, respectively, which are way below the acceptable value ($\sim$74.9\%), while our solution delivers a $\sim$73\% accuracy. Note that our training is at a global batch size of 16128, and it is well known \cite{you2017large, you2019large} that the accuracy drops with the large-batch training due to the generalization gap. The application of layer-wise adaptive rate scaling (LARS) to the decentralized setting might be an option to further improve the performance of our approach. We propose the exploration on LARS in decentralized learning as our future work.     

\section{Related Work}
\label{sec:related}

{\bf White-box Analysis}. White-box analysis is considered as an important approach to understand and refine the DL training process.
DeepXplore~\cite{pei2017deepxplore} deploys whitebox analysis to test real-world DL systems. It is able to find and fix unexpected behavior of a DL model to improve the model accuracy. Pretzel~\cite{lee2018pretzel} treated the entire training pipeline as a whitebox to apply runtime optimizations. Meanwhile, Nasr \ea~\cite{nasr2019comprehensive} also showed that blindly exposing model updating information as a whitebox could reveal privacy vulnerabilities to DL frameworks. In our study, we would assume all workers are trusted and there is no information leakage to malicious parties.


\vspace{2mm}
\noindent {\bf Federated Learning}. Federated learning is a relatively new field of distributed learning. The focus of federated learning is to train a model with data distributed among entities not trusting each other~\cite{kairouz2019advances}. Although federated learning can also benefit from improving communication efficiency~\cite{elgabli2020fgadmm,chai2020fedat} and achieve comparable performance as classic DL~\cite{hegedHus2021decentralized}, its main problem not fully resolved is still ensuring privacy~\cite{mothukuri2021survey}. These efforts to guarantee privacy are orthogonal to our work. The attack model with malicious attackers or Byzantine failures is beyond the scope of this paper.

\vspace{2mm}
\noindent {\bf Optimization on Gradient Norm}. The DL community has already noticed and recognized the optimizations over gradient norms. Gradient clipping~\cite{zhang2019gradient} boosts the training performance regarding to convergence time. Moreover, applying  gradient normalization improves model's explainability~\cite{ross2017neural},  robustness~\cite{ross2018improving}, generalization~\cite{you2019large} and scalability~\cite{you2017large}. GradNorm~\cite{chen2018gradnorm} introduced gradient norm to for training deep multitask networks. These approaches based on gradient norm could be regarded as special cases of gradient compression but it actively affects gradient propagation instead of passively reducing communication overheads only. Our work further sheds light on the potential of gradient norm to improve DL training performance.

\section{Conclusion}
This work introduces a benchmarking framework and methodology to uncover the correlations between model accuracy and variances of parameter tensors across communication graphs and training scales. 
We observed the characteristics of decentralized learning and applied the observations on a decentralized adaptive SGD method. To evaluate the performance of our solutions, we applied the approach   
to train four sample applications at large scales.   
 The results show that our approach obtains the best convergence rates in decentralized DNN training, and delivers equally or comparably good model accuracy for all applications as centralized learning does. 

\newpage


\nocite{langley00}

\bibliography{ref.bib}
\bibliographystyle{mlsys2023}


\end{document}